\documentclass[preprint,12pt]{elsarticle}

\usepackage{amssymb}
\usepackage{amsmath}
\usepackage{xcolor}
\usepackage{float}
\usepackage{graphicx}
\usepackage{hyperref}
\usepackage{multirow}
\usepackage{graphicx}
\journal{XXX}

\begin{document}

\begin{frontmatter}



\title{A Comparative Study on Dynamic Graph Embedding based on Mamba and Transformers}

\author[1]{Ashish Parmanand Pandey}
\address[1]{Department of Data Science, New Jersey Institute of Technology, Newark, NJ 07102, USA}
%\affiliation[label1]{organization={},
\author[2]{Alan John Varghese}
\address[2]{School of Engineering, Brown University, Providence, RI 02912, USA}

\author[1]{Sarang Patil}

\author[1]{Mengjia Xu \corref{Corresponding author}}


%\author{} %% Author name
\fntext[Corresponding author]{Corresponding author: mx6@njit.edu}
%\affiliation{organization={},%Department and Organization

%% Abstract
\begin{abstract}
Dynamic graph embedding has emerged as an important technique for modeling complex time-evolving networks across diverse domains. While transformer-based models have shown promise in capturing long-range dependencies in temporal graph data, they face scalability challenges due to quadratic computational complexity. This study presents a comparative analysis of dynamic graph embedding approaches using transformers and the recently proposed Mamba architecture, a state-space model with linear complexity. We introduce three novel models: TransformerG2G augmented with graph convolutional networks, $\mathcal{DG}$-Mamba, and $\mathcal{GDG}$-Mamba with graph isomorphism network edge convolutions. Our experiments on multiple benchmark datasets demonstrate that Mamba-based models achieve comparable or superior performance to transformer-based approaches in link prediction tasks while offering significant computational efficiency gains on longer sequences. Notably, $\mathcal{DG}$-Mamba variants consistently outperform transformer-based models on datasets with high temporal variability, such as UCI, Bitcoin-OTC, and Reality Mining, while maintaining competitive performance on more stable graphs like SBM. We provide insights into the learned temporal dependencies through analysis of attention weights and state matrices, revealing the models' ability to capture complex temporal patterns. By effectively combining state-space models with graph neural networks, our work addresses key limitations of previous approaches and contributes to the growing body of research on efficient temporal graph representation learning. These findings offer promising directions for scaling dynamic graph embedding to larger, more complex real-world networks, potentially enabling new applications in areas such as social network analysis, financial modeling, and biological system dynamics.
\end{abstract}



%% Keywords
\begin{keyword}
graph embedding \sep transformers \sep dynamic graphs \sep link prediction \sep state-space models \sep long-term dependencies


\end{keyword}

\end{frontmatter}


\section{Introduction}

Studying the evolving dynamics exhibited in complex real-world graphs (or networks) across spatial and temporal scales plays a key role in diverse fields such as biological networks, social networks, e-commerce, transportation systems, functional brain connectomes, and financial markets. We can model these graphs as dynamic graphs in either continuous time or discrete time, based on the nature of the data. Understanding how these networks change over time and space can provide great insight into their underlying structures and behaviors, enabling more accurate future forecasting in areas such as disease progression modeling, transportation planning, anomaly detection, and risk assessment. Conventional graph neural networks (GNNs)~\cite{kipf2016semi,  gilmer2017neural,velivckovic2017graph,yun2019graph} assume a fixed graph structure without considering evolving dynamics. Similarly, advanced static graph embedding methods, such as those preserving discriminative and geometric properties~\cite{gou2023discriminative}, are limited to static structures and do not account for temporal evolution.
However, time-aware temporal graph embedding techniques, including continuous-time~\cite{trivedi2019dyrep,kumar2019predicting,rossi2020temporal} and discrete-time methods~\cite{goyal2018dyngem, goyal2018dynamicgem, pareja2020evolvegcn, xu2022dyng2g, you2022roland, sharma2023temporal}, facilitate learning temporal information for future forecasting by integrating conventional GNNs with recurrent neural networks (RNNs) or long short-term memory (LSTM) or gated recurrent unit (GRU). However, RNN or LSTM-based temporal graph embedding models often suffer from issues like catastrophic forgetting, vanishing gradients, and exploding gradients. Moreover, these approaches face significant challenges while dealing with very large sequential graph snapshots. 

On the other hand, attention-based transformers~\cite{vaswani2017attention} have empowered many advanced Large Language Models (LLMs) with impressive in-context learning (ICL) capability. Transformers have been increasingly applied to capture long-range dependencies across graph snapshots, enabling parallel processing of long graph sequences and significantly enhancing context awareness. Dynamic graph representation learning applications have made impressive progress with the development of attention mechanisms for diverse applications, such as learning temporal graph embeddings with uncertainty quantification~\cite{varghese2024transformerg2g}, anomaly detection~\cite{liu2021anomaly}, traffic flow forecasting~\cite{huo2023hierarchical, jiang2023pdformer}, financial time series prediction~\cite{lezmi2023time}, weather forecasting~\cite{wang2024dynamic}, heating environment forecasting~\cite{hofmeister2024dynamic}, and 3D object detection~\cite{ren2023dynamic}. Diverse PyTorch implementations of the aforementioned temporal graph embedding methods can be found in  \url{https://pytorch-geometric-temporal.readthedocs.io/}. Nevertheless, the quadratic computational complexity of transformers when dealing with long graph snapshots remains a major shortcoming despite their promising performance and distinct advantage in capturing long-range dependency. 

More recently, an emerging alternative to transformers, known as ``state-space sequence models (SSMs)''~\cite{gu2021efficiently, gu2023mamba}, has gained substantial attention by addressing the ``quadratic complexity'' bottleneck inherent in the attention mechanism. In particular, a selective SSM architecture, Mamba~\cite{gu2023mamba}, achieved impressive performance on the Long Range Arena (LRA) benchmark with only linear complexity. The Mamba model outperformed the initially developed structured SSM model (i.e., S4 model~\cite{gu2021efficiently}) by incorporating the ``selective scan'' technique to compress the data selectively into the state and improved efficiency through the adoption of hardware-aware algorithms on modern hardware (e.g., GPUs and TPUs). Due to temporal graphs that can be modeled as continuous-time temporal events or discrete-time graph snapshots~\cite{longa2023graph}, Li et al.~\cite{li2024state} developed an SSM-based framework (i.e., GraphSSM) for learning temporal graph dynamics by integrating structural information into the online approximation objective with a Laplacian regularization term. 

Herein, we conducted a comparative study on dynamic graph embedding based on Mamba and Transformers for diverse benchmarks with smooth and heterogeneous temporal dynamics.  

\section{Preliminaries}

\subsection{Dynamic graph modeling}
Dynamic graphs provide a powerful framework for representing complex systems that evolve over time in various domains \cite{holme2012temporal,kazemi2020representation}. Unlike static graphs, dynamic graphs capture the temporal evolution of relationships between entities, offering a more realistic model of real-world networks~\cite{xu2020inductive}. Generally, a discrete-time dynamic graph $G$ can be represented by a sequence of graph snapshots $G = \{G_1, G_2, ..., G_T\}$, 
where $G_t = (V_t, E_t), t \in [0,T]$ denotes the state of the graph at time step $t$, where $V_t$ is the set of nodes and $E_t$ is the set of edges~\cite{xu2020inductive}. More precisely, 
\begin{equation}
\begin{aligned}
V_t &= \{v_1, v_2, ..., v_{|V_t|}\}, \; E_t \subseteq \{(u, v) | u, v \in V_t\}.
\end{aligned}
\end{equation}

The discrete-time representation allows us to model systems where changes occur at specific intervals, such as daily social media interactions or weekly financial transactions~\cite{singer2019node}.

In practice, the set of nodes $V_t$ can change over time, reflecting the dynamic nature of entity participation in the system. These dynamics can be represented as $V_i \neq V_j$ for time steps $i \neq j$, and $i, j \in [0, T]$. The corresponding adjacency matrix $A_t \in \mathbb{R}^{|V_t|\times |V_t|}$ of graph $G_t$ at time $t$ can be defined as

\begin{equation}
A_t = [a_t (u,v)], \text{ where } a_t (u,v) = 
\begin{cases}
1  \quad \text{or} \quad w_{uv} & \text{if } (u, v) \in E_t; \\
0 & \text{otherwise}.
\end{cases}
\end{equation}
Here, $w_{uv}$ denotes the weight associated with the edge between node pair ($u,v$) when the graph $G_t$ is weighted; Otherwise, in an unweighted graph, the edge weight is uniformly $1$.
The key challenge in dynamic graph modeling lies in effectively and accurately capturing both the spatial and temporal dependencies. The primary goal of temporal graph embedding is to learn a nonlinear mapping function $f$ that maps each node $v$ at time $t$ to a low-dimensional embedding vector:
\begin{equation}
f : (v, t) \rightarrow emb(v,t) \in \mathbb{R}^{L_o}, v\in V_t
\end{equation}
where $L_o$ is the dimensionality of the node embedding vector, typically much smaller than $|V_t|$.

To address the above challenges, sequence models such as RNNs and their variants (LSTMs and GRUs), as well as transformers, have demonstrated effectiveness in capturing temporal dynamics across various domains, e.g., machine translations, sentiment analysis, and time series analysis. However, each technique has its own benefits and limitations. For example, RNNs model the temporal evolution via nonlinear recurrence using the following equations \cite{seo2018structured}.

\begin{equation}
\begin{aligned}\label{eq:RNNs}
h_t &= \sigma(W_xx_t + W_hh_{t-1} + b_h),\\
y_t &= \sigma(W_oh_{t} + b_o),
\end{aligned}
\end{equation}
where $h_t$ is the hidden state at time $t$, $x_t$ is the input feature vector, $y_t$ is the output, $W_x, W_h,$ and $W_o$ are weight matrices, $b_h$ and $b_o$ are bias vectors, and $\sigma$ is a nonlinear activation function (i.e., tanh). LSTMs and GRUs apply different types of gates to mitigate the vanishing gradient issue encountered in conventional RNNs. Specifically, gates help control the information flow, enabling the model to retain important long-term dependencies while discarding irrelevant information. However, RNNs and their variants process tokens sequentially and rely on backpropagation through time (BPTT) for optimization, which limits their scalability and efficiency when handling long sequences. Moreover, recurrent model architectures are prone to a critical issue, i.e., catastrophic forgetting.

{\it Attention-based Transformer models}, on the other hand, enable to compute the ``contextually rich'' embedding of a node $z$ at time $t$ as a weighted sum of its historical representations:
\begin{equation}
emb(z,t) = \sum_{i=1}^{t} \alpha_i h_i, 
\end{equation}
where $\alpha_i$ are attention weights that quantify the relevance of each historical representation to the current time step $T$. These weights are computed on the basis of the relationships between past states and the current prediction, allowing the model to focus more on the most informative parts of the history and capture long-term dependencies better than RNNs. However, computing pairwise attention scores can lead to quadratic complexity in terms of the sequence length $L$, resulting in a significant computational cost of $\mathcal{O}(L^2)$ and a high memory demand when $L$ is very large. 
% Some recent works also focus on enhancing the efficiency of Transformers~\cite{katharopoulos2020transformers}.

\subsection{State space models (SSMs)}

Recently, control theory-inspired State-Space Models (SSMs), an emerging alternative to transformers, offer a more general technique with linear complexity that outperforms transformers, particularly in long-sequence modeling tasks (e.g., the Google LRA benchmark\footnote{https://github.com/google-research/long-range-arena}). Due to their advantages, deep SSMs have been increasingly applied in various domains, such as speech recognition, language processing, time series prediction, music generation, computer vision, temporal graph modeling, and bioinformatics~\cite{patro2024mamba}. Mathematically, SSMs model a system's internal state and its temporal evolution through a recursive state transition equation, which enables capturing long-range dependencies and modeling complex dynamics, without the need for attention mechanisms. SSMs typically contain two main types of representation: continuous-time and discrete-time. Continuous SSMs can be simply described as linear ordinary differential equations~\cite{gu2021efficiently}, while discrete SSMs can be modeled in both recurrent and convolutional forms. The recurrent form of discrete SSMs is shown in Eq.~\ref{eq:ssm_rec_form}. It is equivalent to an RNN without nonlinear recurrent activations. 
\begin{equation}
\begin{aligned}
\label{eq:ssm_rec_form}
\text{state equation:} &\quad h_{t}= \bar{A}h_{t-1} + \bar{B}x_t \\
\text{output equation:} &\quad y_t = \bar{C}h_t + \bar{D}x_t
\end{aligned}
\end{equation}
where $x_t \in \mathbb{R}^{d_{in}}$ is the input sequence, $y_t \in \mathbb{R}^{d_{out}}$ is the output sequence, $h_t \in \mathbb{R}^m$ is the state vector, $\bar{A} \in \mathbb{R}^{m\times m}$ is the state transition matrix, $\bar{B} \in \mathbb{R}^{m\times d_{in}}$ is the input matrix, $\bar{C} \in \mathbb{R}^{d_{out} \times m}$ is the output matrix, and $\bar{D} \in \mathbb{R}^{d_{out} \times d_{in}}$ is the skip connection matrix. $\bar{A}, \bar{B}, \bar{C}$ and $\bar{D}$ are learnable SSM parameters. In the context of dynamic graph embedding, $\mathbf{h}_t$ represents the latent node or edge embedding of the graph, $\mathbf{x}_t$ could be new node or edge features, and $\mathbf{y}_t$ is the predicted label (classification problem) or value (regression problem) at time $t$.

Discrete-time SSMs can also be formulated as a global convolution operation in terms of an SSM convolutional kernel, \( \bar{K} \), which has the same shape as the input sequence length $L$; see the detailed formula in Eq.~\ref{eq:ssm_conv}. The SSM convolutional kernel is composed of the three SSM parameters ($\bar{A}, \bar{B}$, and $\bar{C}$). The Convolutional representation of SSMs offers several advantages, including computational efficiency through the use of fast algorithms like Fast Fourier Transform (FFT):
\begin{equation}
\label{eq:ssm_conv}
y_t = \bar{K}*x_t, \quad \text{where} \quad \bar{K} \in \mathbb{R}^L: (\bar{C}\bar{A}^k\bar{B})_{k\in[L]}.
\end{equation}

The structured SSMs (S4) were initially developed as linear time invariant (LTI) models, with all parameters ($\bar{A},\bar{B}, \bar{C}$, and $\Delta$) fixed across time steps for the entire sequence. Here, $\Delta$ represents the discretization step size, which determines the temporal resolution of the state-space model by converting continuous-time dynamics to discrete-time dynamics~\cite{gu2021efficiently}. In particular, the S4 model parameterized the state matrix ($A$) as a HIPPO (High-order Polynomial Projection Operator) matrix~\cite{gu2021efficiently}, which allows it to effectively capture the underlying temporal dynamics by projecting discrete time series onto polynomial bases. However, the fixed parameterization approach restricts the effectiveness of S4 in content-based reasoning tasks. To address this issue, the Selective SSMs, also known as S6 or Mamba~\cite{gu2023mamba}, incorporate the selectivity mechanism that updates the state selectively based on the input, i.e., $\bar{B} = S_B(x_t), \bar{C} = S_C(x_t), \Delta_t = S_{\Delta_t} (\Delta_t)$ are now input-dependent functions, allowing selective update of the hidden state based on the relevance of incoming inputs. Hence, Mamba is also known as a linear time-variant (LTV) SSM. Moreover, Mamba employs hardware-aware algorithms, including parallel scan, kernel fusion, and activation recomputation, to achieve greater efficiency when processing long sequences. 

The aforementioned advances in SSMs offer promising directions for modeling dynamic graphs, potentially providing more efficient and effective alternatives to attention-based models to capture long-range dependencies in temporal graphs \cite{li2024state}. Each of these approaches offers unique trade-offs in terms of expressiveness, computational efficiency, and the ability to capture long-term dependencies. In this paper, we will explore how various architectures, including transformers and state-space models, can be adapted to address the dynamic graph embedding problem. Our focus will be on their capacity to capture complex spatial-temporal patterns and effectively scale to large, evolving networks. To address the aforementioned challenges, we have introduced three new probabilistic temporal graph embedding models based on our prior works~\cite{varghese2024transformerg2g, xu2022dyng2g}, {\it i.e., ST-transformerG2G, $\mathcal{DG}$-Mamba, and $\mathcal{DG}$-Mamba with GINE Convolutions}. These models effectively integrate spatial and temporal dependencies as well as incorporate important evolving edge features, using transformers and emerging Mamba models to enhance performance in temporal link prediction.

%%%%%%%%%%%%%%%%%%%%%%%%%%%%%%%%%%%%%%%%%%%%%
\section{Proposed Methods} 
\label{methodology}
%%%%%%%%%%%%%%%%%%%%%%%%%%%%%%%%%%%%%%%%%%%%%

\paragraph{Problem Definition} We consider a discrete-time temporal graph $\mathcal{G} = \{G_t\}_{t=1}^T$, represented by a sequence of graph snapshots over $T$ timestamps. Each graph snapshot $G_t$ consists of a vertex set $V_t = \{ v_1^t, v_2^t, v_3^t, \ldots, v_{|V_t|}^t \}, t \in [T]$ and an edge set $E_t = \{e_{i,j}^t | i^t, j^t \in V_t\}$, where each edge $e_{i,j}^t$ represent a connection between nodes $i^t$ and $j^t$. Each node $v_i^t$ will have a set of node representations matrix $X_t = \{x_i^t| i = 1, 2, ..., |V_t|\}$. The main objective here is to transform the representation of nodes from a high-dimensional sparse non-Euclidean space to a lower-dimensional dense space as multivariate Gaussian distributions at each timestamp. Representing each node as a Gaussian distribution allows us to capture the uncertainty associated with that representation. To this end, our aim is to learn node representations for the timestamp $t$ using graph snapshots from the current and the previous $l$ timestamps, i.e., $G_{t - l}, \ldots, G_{t-1}, G_t$, while capturing the long-range dependencies at both spatial and temporal scales. Here, the look-back $l$ is a predefined hyperparameter that represents the number of timestamps to look-back, i.e., the length of the history considered.

\paragraph{Probabilistic Node Representation} To achieve the main aims described above, three temporal graph embedding models were developed in the paper. Specifically, all proposed models learn a non-linear mapping function that encodes every node in a graph snapshot in the graph sequence $\mathcal{G}$ into a lower-dimensional embedding space as probabilistic density functions, i.e., 
each node is represented as a 
multivariate Gaussian distribution $\mathcal{N}(\mu_i^t, \Sigma_i^t)$, where $\mu_i^t\in \mathcal{R}^{L_o}$ denotes the embedding means; $\Sigma_i^t \in \mathcal{R}^{L_o\times L_o} = diag(\sigma_i^t)$ represents the square diagonal covariance matrix with the embedding variances on the diagonal, $L_o$ is the size of node embedding.

\paragraph{Data Preprocessing} Due to the varying sizes of input graph snapshots across time in the temporal graph, all adjacency matrices ($\{A_t\}_{t=1}^T$) in the graph sequence are initially normalized to adjacency matrices of same size ($\{\tilde{A}_t\}_{t=1}^T)$ with zero-padding technique. As such, all the normalized adjacency matrices have the same shape of $(n, n)$, where $n$ represents the maximum number of nodes across all timestamps. If a node is absent in the graph snapshot at a particular timestamp $t$, the corresponding row and column in the normalized adjacency matrix $\tilde{A}_t$ are filled with zeroes, indicating that the node has no connections to any other nodes during that timestamp.

\subsection{ST-TransformerG2G: Improved TransformerG2G with GCN model}

\begin{figure}[!h]
    \centering
    \includegraphics[width = 0.7\textwidth]{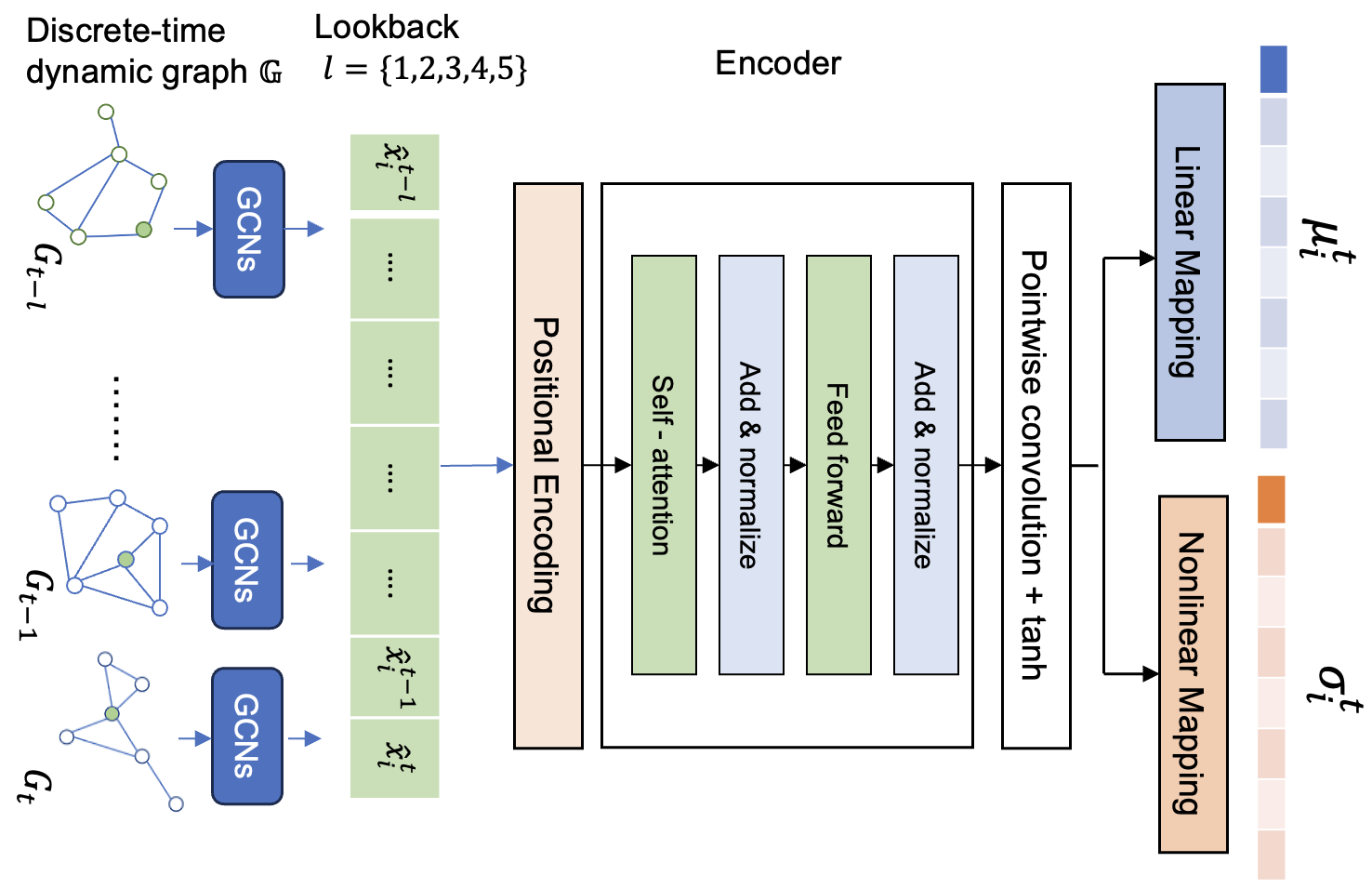}
    \caption{The proposed ST-TransformerG2G model architecture enhanced the previous TransformerG2G model~\cite{varghese2024transformerg2g} with GCNs. An additional GCN block, consisting of three GCN layers, was added into the TransformerG2G model, to explicitly capture the spatial interactions for each graph snapshot. The generated node embeddings were fed into a vanilla transformer encoder module, with positional encoding added to each input node token embedding.The final output representation of each node is a multivariate normal distribution - $\mathcal{N}(\mu_i^t, \Sigma_i^t)$, where $\Sigma_i^t = diag(\sigma_i^t)$.}
    \label{fig:ST_transformerG2G}
\end{figure}

In this model, we build on our previous TransformerG2G model \cite{varghese2024transformerg2g} by including GCNs blocks to explicitly capture spatial dependencies within each graph snapshot, see detailed architecture in Fig.~\ref{fig:ST_transformerG2G}. The model takes as input a series of graphs $(G_{t-l}, \ldots, G_t)$, and each graph is passed through GCN blocks. The GCN block consists of 3 GCN layers that help in learning spatial information. From the output of the GCNs block, we obtain a sequence of learned representation of node $i$ over $l+1$ timestamps: $(\hat{x}_i^{t-l}, \cdots, \hat{x}_i^{t-1}, \hat{x}_i^t)$, where $\hat{x}_i^t \in \mathbb{R}^d $ and $\hat{x}_i^t$ corresponds to node $i$'s representation at timestamp $t$. We also added a vanilla positional encoding to this sequence of vectors~\citep{vaswani2017attention}. The resulting sequence is then processed by a standard transformer encoder block \cite{vaswani2017attention} with single-head attention, which outputs another sequence of $d$-dimensional vectors. The sequence of output vectors from the encoder block is combined using a ``point-wise convolution'' to obtain a single vector, and a tanh activation is applied to obtain $h_i$. This vector is then fed through two projection heads to obtain the final probabilistic node embedding means -- $\mu_i^t \in \mathbb{R}^{L_o}$ and variance -- $\sigma_i^t \in \mathbb{R}^{L_o}$, representing the node $v_i^t$. The projection head to obtain the mean vector consists of a linear layer, i.e., $\mu_i = h_iW_{\mu} + b_{\mu}$, and the projection head to obtain the variance vector consists of a linear layer with $elu$ activation, i.e., $\sigma_i = elu(h_i W_{\Sigma} + b_{\Sigma}) + 1,$ to ensure the positivity of the variance vector. The self-attention mechanism in the encoder block enables the model to learn the information from the node's historical context, and the GCN block helps learn the spatial dependencies in each graph.

% ###### Model 2:
\subsection{$\mathcal{DG}$-Mamba model}

The $\mathcal{DG}$-Mamba model offers an alternative to our earlier TransformerG2G architecture~\cite{varghese2024transformerg2g} for temporal graph embedding. It utilizes selective scan-based SSMs (i.e., Mamba)~\cite{gu2023mamba} to efficiently capture long-range dependencies without the attention mechanism. The newly proposed $\mathcal{DG}$-Mamba framework makes use of Mamba's enhanced efficiency to overcome the quadratic autoregressive inference complexity bottleneck of transformers, while also effectively managing long-range dependencies in dynamic graphs.

\begin{figure}[!h]
    \centering
    \includegraphics[width=.8\linewidth]{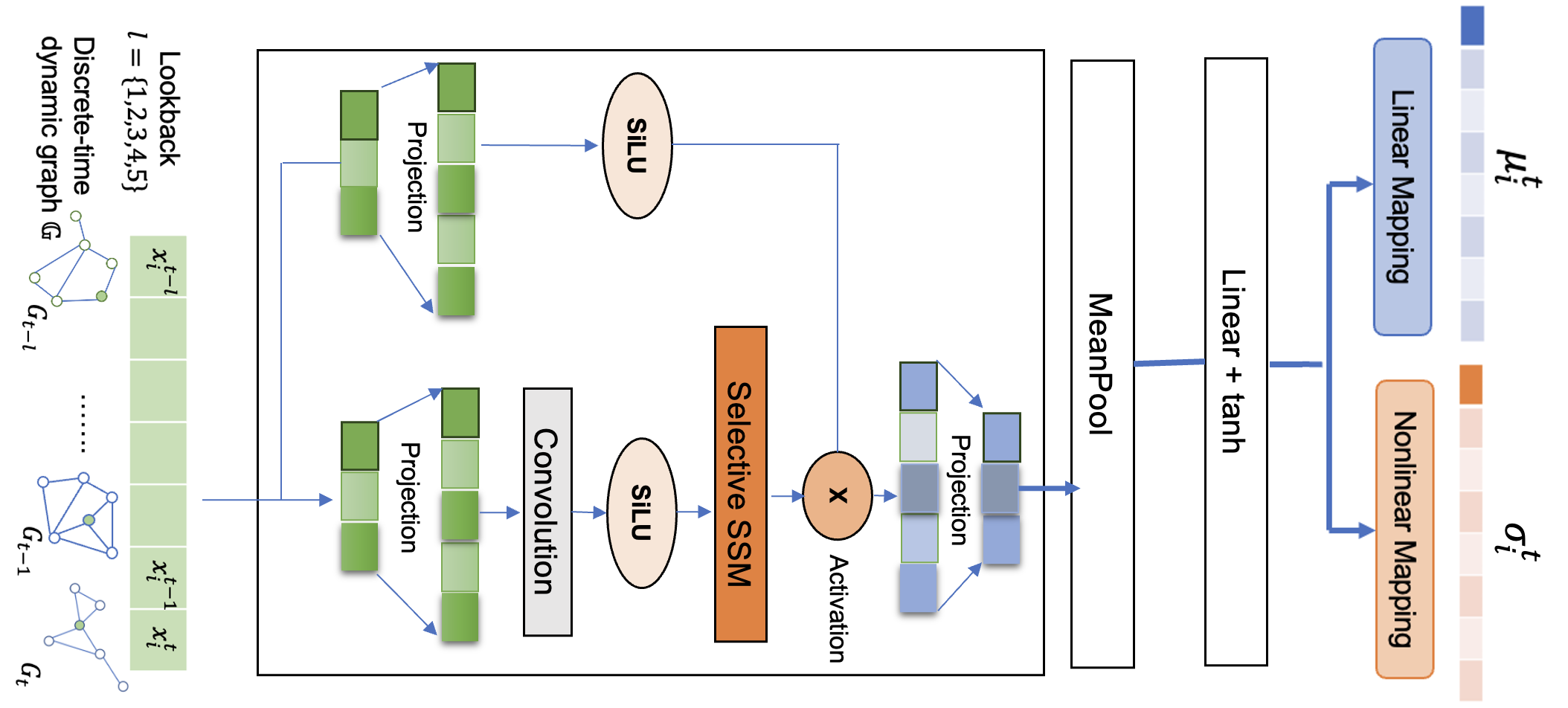} 
    \caption{The main architecture of the proposed $\mathcal{DG}$-Mamba model. It processes a sequence of discrete-time graph snapshots $\{G_t\}_{t=1}^T$, where a look-back parameter $l = \{1, 2, 3, 4, 5\}$ allows for historical context integration. Each graph snapshot first goes through projection and convolution to capture localized node features while maintaining spatial relationships. The SSM layer is followed to efficiently capture long-range temporal dependencies using the selective scan mechanism (Mamba architecture). The output of this Mamba layer is then passed through the activation function for non-linearity, followed by mean pooling to generate an aggregate representation. A linear projection layer with tanh activation is used to refine the node embeddings. An additional two projection heads (including one linear projection layer and one nonlinear projection mapping with \texttt{ELU} activation function) are used to obtain the mean and variance of Gaussian embeddings.}
    \label{fig:TG-mamba}
\end{figure}

The main architecture of $\mathcal{DG}$-Mamba is illustrated in Fig.~\ref{fig:TG-mamba}. It follows four main steps: a) data preprocessing, b) temporal graph embedding based on Mamba, c) embedding aggregation, and d) output probabilistic distributions. Initially, it processes discrete-time input graph snapshots using the aforementioned zero-padding technique to standardize their sizes. In order to capture long-range dependencies for each graph node, the history length parameter, i.e., look-back $l$, is set to $l = \{1,2,3,4,5\}$. Each graph snapshot $G_t = (A_t, X_t)$ has a corresponding adjacency matrix $A_t$ and a node feature matrix $X_t$, it follows the following procedure to obtain the node-level Gaussian embeddings.

The core operation starts with the Mamba Layer. The input to this layer is processed using a selective scan-based state-space model (SSM)~\cite{gu2023mamba}.The input graph node feature $x_i$ at each timestamp $t$ is processed using the Mamba layer and outputs embedding $e_i$:
\begin{equation}
e_i = \text{Mamba}(x_i, l),
\end{equation}
where $x_i = \{x_i^t\}_{t=1}^{T}$ is the input sequence of node features for node $i$, and $l$ is the lookback. Next, the embedding $e_i$ is passed through the Mean Pooling layer for aggregating the graph representation:
\begin{equation}
e_i = \text{MeanPool}(e_i).
\end{equation}

The pooled embedding is further transformed through a linear layer followed by the \texttt{tanh} activation function.
\begin{equation}
h_i = \tanh(\text{Linear}(e_i)). 
\end{equation}

The nonlinear exponential linear unit (ELU) activation function is added to capture richer features: $h_i = \text{ELU}(h_i)$. The processed embedding is then parameterized as a Gaussian distribution by computing the mean and variance as follows.
\begin{equation}
\begin{aligned}
\mu_i &= \text{Linear}_\mu(h_i), \\
\sigma_i &= \text{ELU}(\text{Linear}_\sigma(h_i)) + 1 + \epsilon.
\end{aligned}
\end{equation}
Here, $\epsilon$ is a small constant (1e-14) added for numerical stability.

The $\mathcal{DG}$-Mamba model preserves the probabilistic embedding characteristic of TransformerG2G~\cite{varghese2024transformerg2g}, representing each node as a multivariate Gaussian distribution in the latent space. Furthermore, it enhances the previous model's ability to capture both spatial and temporal dynamics. This allows for uncertainty quantification in the learned representations, which is particularly valuable for dynamic graphs where node properties and relationships may change over time. 

% ###### Model 3: 
\subsection{$\mathcal{GDG}$-Mamba Model: Enhanced $\mathcal{DG}$-Mamba with GINE Conv}

\begin{figure}[!h]
    \centering
    \includegraphics[width=.75\linewidth]{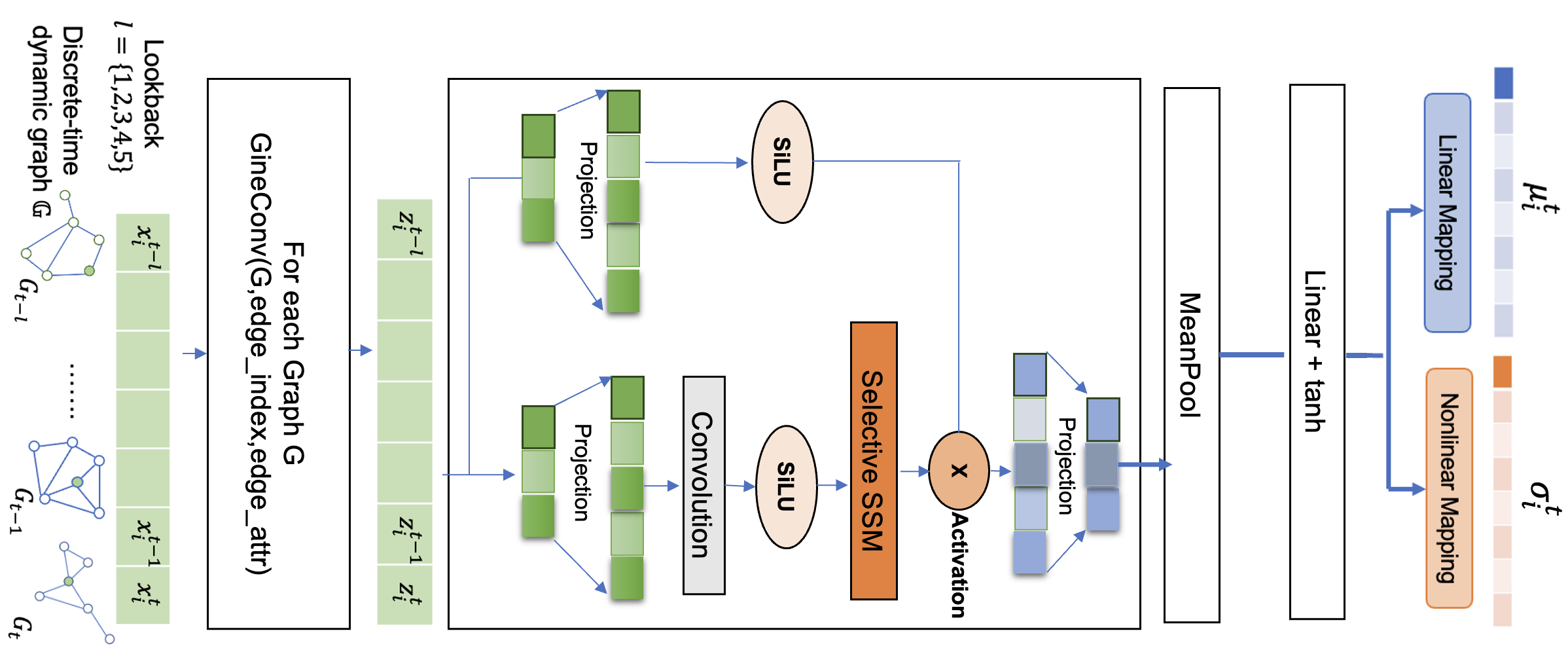}
    \caption{The main architecture of the $\mathcal{GDG}$-Mamba model. It first processes a series of discrete-time graph snapshots $\{G_t\}_{t=1}^T$ using the GINE convolution to enhance the spatial representation of the graph by considering both node and edge-level features at each timestamp. The generated graph sequence representations are then processed through the Mamba block to capture temporal dynamics, followed by mean pooling and a linear layer with tanh nonlinearity before outputting the final Gaussian embeddings. The model incorporates node and edge-level features across both spatial and temporal dimensions.}
    \label{fig:DG_mamba_GINEConv}
\end{figure}

To enhance spatial representation by incorporating node features and crucial edge features simultaneously, before learning temporal dynamics with the Mamba layer, we introduce a $\mathcal{DG}$-Mamba variant ($\mathcal{GDG}$-Mamba) based on the Graph Isomorphism Network Edge (GINE) convolutions \cite{behrouz2024graph}; the detailed architecture is shown in Fig.~\ref{fig:DG_mamba_GINEConv}.

The main procedure of this enhanced $\mathcal{DG}$-Mamba model involves the following steps. First, the input graph node feature $x_i^t, t\in[T]$ at each timestamp $t$ is processed using Graph Isomorphism Network Edge (GINE) convolution to enhance spatial representation.
\begin{equation}z_i^t = \text{GINEConv}(x_i^t, \text{edge\_index}, \text{edge\_attr})
\end{equation}
The \texttt{GINEConv} operation captures the structural information and generates an enriched node representation $z_i^t$ by considering both node and edge features. The processed node representations for all timestamps are concatenated $z_i = \{z_i^t|t = 1, 2, ..., T\}$. Then this sequence is passed through the Mamba layer, followed by mean pooling, linear projection with \texttt{tanh} activation function to generate an intermediate representation for each node. Additional linear layers for mean and log-variance of Gaussian embeddings as shown in Eq. \ref{eqn:tgmamba}. In the same way we processed $\mathcal{DG}$-Mamba earlier, except here we pass the GINE Convolution processed output $z_i$ instead of directly passing $x_i$. 
\begin{equation}
\label{eqn:tgmamba}
\begin{aligned}
e_i &= \text{Mamba}(z_i, l) \\
e_i &= \text{MeanPool}(e_i) \\
h_i &= \tanh(\text{Linear}(e_i)) \\
h_i &= \text{ELU}(h_i) \\
\mu_i &= \text{Linear}_\mu(h_i) \\
\sigma_i &= \text{ELU}(\text{Linear}_\sigma(h_i)) + 1 + \epsilon
\end{aligned}
\end{equation}

The GINE convolution is defined as:
\vspace{-0.2cm}
\fontsize{9}{9.5}
\begin{equation}
\text{GINEConv}(x_i^t, \text{edge\_index}, \text{edge\_attr}) = \phi((1 + \epsilon) \cdot x_i^t + \sum_{j \in \mathcal{N}(i)} \text{ReLU}(x_j^t \| \text{edge\_attr}_{(i,j)}))
\end{equation}
\normalsize

where $\phi$ is a learnable function (typically an MLP), $\epsilon$ is a learnable parameter, $\mathcal{N}(i)$ denotes the neighbors of node $i$, and $\|$ represents concatenation.

The proposed models (ST-TransformerG2G, $\mathcal{DG}$-Mamba, and enhanced $\mathcal{GDG}$-Mamba) apply the same triplet-based contrastive loss for training, encouraging similar nodes to have close Gaussian distributions in the embedding space while pushing dissimilar nodes apart.

\subsection{Loss function and training methodology}

All models considered in this work employ a triplet-based contrastive loss, inspired by \cite{bojchevski2018deep}, to train the network. The triplet-based contrastive loss is defined in Eq. \ref{eq:triplet_loss}. In this process, each node acts as a reference point, for which we sample the $k$-hop neighborhoods - nodes that are atmost $k$ hops away from the reference node. Using these neighborhoods, we construct a set of node triplets $\mathcal{T}_t = \{ (v_i, v_i^{near}, v_i^{far})| v_i \in V_t\}$ for each timestamp. Each triplet consists of a reference node $v_i$, a nearby node $v_i^{near}$, and a distant node $v_i^{far}$. These triplets satisfy the constraint that the shortest path from $v_i$ to $v_i^{near}$ is shorter than that to $v_i^{far}$, i.e., $sp(v_i, v_i^{near}) < sp(v_i, v_i^{far})$. The function $sp(.,.)$ measures the shortest path length between two nodes. The triplet-based contrastive loss is defined as follows:
\begin{equation}
\label{eq:triplet_loss}
    \mathcal{L} = \sum_t \sum_{\left(v_i, v_i^{near}, v_i^{far}\right) \in \mathcal{T}_t} \left[ \mathbb{E}^2_{\left(v_i, v_i^{near}\right)}  + e^{-\mathbb{E}_{\left(v_i, v_i^{far}\right)}}\right],
\end{equation}
where $\mathbb{E}_{\left(v_i, v_i^{near}\right)}$ and $\mathbb{E}_{\left(v_i, v_i^{far}\right)}$ represent the Kullback-Leibler (KL) divergence between the embeddings of the reference node and its nearby and distant nodes, respectively. The KL divergence quantifies the dissimilarity between the joint normal distributions of two nodes in the embedding space. The formula for calculating the KL divergence between the multivariate Gaussian embeddings of two nodes ($v_i$ and $v_j$) is shown in Eq.~\ref{eq: kl_divergence}.
\begin{equation}
\label{eq: kl_divergence}    
\begin{split}
    \mathbb{E}_{\left( v_i, v_j \right)} &= D_{KL}\left( \mathcal{N}(\mu_i,\Sigma_i), \| \mathcal{N}(\mu_j, \Sigma_j)  \right)\\
    &= \frac{1}{2} \left[ tr(\Sigma_j^{-1} \Sigma_i) + (\mu_j - \mu_i)^T \Sigma_j^{-1}(\mu_j-\mu_i) - L + log\frac{|\Sigma_j|}{|\Sigma_i|}\right].
\end{split}
\end{equation}
The triplet-based contrastive loss aims to minimize the dissimilarity between the distributions of $v_i$ and $v_i^{near}$, and maximize the dissimilarity between the distributions of $v_i$ and $v_i^{far}$.

\section{Experimental Results and Analysis} 
\label{results}

\subsection{Dataset descriptions}

We used five different graph benchmarks to validate and compare our proposed models. A summary of the dataset statistics -- including {\it the number of nodes, number of edges, number of timesteps, and the distribution of train/validation/test splits with specific numbers of timesteps, as well as the embedding size} -- is provided in Table~\ref{tab:benchmark_table}.

\vspace{-.2in}
\begin{table}[!ht]
\centering
\caption{Details of the experimental datasets.}
\label{tab:benchmark_table}
\setlength{\tabcolsep}{2.5pt}
\fontsize{9.5pt}{11pt}\selectfont 
\renewcommand{\arraystretch}{1.3} % Increase row height by 30%
\begin{tabular}{l||lllll}
     \hline
     Dataset & \#Nodes & \#Edges & \#Timestamps   & \#Train/Val/Test & Embedding Size ($L_o$)\\ 
     \hline \hline
     Reality Mining & 96 & 1,086,403 & 90 & 63/9/18 & 64\\
     UCI & 1,899 & 59,835 & 88 & 62/9/17 & 256\\
     SBM & 1,000 & 4,870,863 & 50 & 35/5/10 & 64\\
     Bitcoin-OTC & 5,881 & 35,588 & 137 & 95/14/28 & 256\\ 
     Slashdot & 50,825 & 42,968 & 12 & 8/2/2 & 64\\ \hline
\end{tabular}
\end{table}

% social network
\textbf{Reality Mining dataset\footnote{\url{http://realitycommons.media.mit.edu/realitymining.html}}}: The network contains human contact data among 100 students of the Massachusetts Institute of Technology (MIT); the data was collected with 100 mobile phones over 9 months in 2004. Each node represents a student; an edge denotes the physical contact between two nodes. In our experiment, the dataset contains 96 nodes and 1,086,403 undirected edges across 90 timestamps.

% social network
\textbf{UC Irvine messages (UCI) dataset\footnote{\url{http://konect.cc/networks/opsahl-ucsocial/}}}: It contains sent messages between the users of the online student community at the University of California, Irvine. The UCI dataset contains 1,899 nodes and 59,835 edges across 88 timestamps (directed graph). This dataset exhibits highly transient dynamics.

% synthetic dataset
\textbf{Stochastic Block Model (SBM) dataset\footnote{\url{https://github.com/IBM/EvolveGCN/tree/master/data}}}: It is generated using the Stochastic Block Model (SBM) model. The first snapshot of the dynamic graph is generated to have three equal-sized communities with an in-block probability of 0.2 and a cross-block probability of 0.01. To generate subsequent graphs, it randomly picks 10-20 nodes at each timestep and moves them to another community. The final generated synthetic SBM graph contains 1000 nodes, 4,870,863 edges, and 50 timestamps.

% finance network 
\textbf{Bitcoin-OTC (Bit-OTC) dataset\footnote{\url{http://snap.stanford.edu/data/soc-sign-bitcoin-otc.html}}}: It is a who-trusts-whom network of people who trade using Bitcoin on a platform called Bitcoin OTC. The Bit-OTC dataset contains 5,881 nodes and 35,588 edges across 137 timestamps (weighted directed graph). This dataset exhibits highly transient dynamics.

% social network
\textbf{Slashdot dataset\footnote{\url{http://konect.cc/networks/slashdot-threads/}}}: It is a large-scale social reply network for the technology website Slashdot. Nodes represent users, and edges correspond to users' replies. The edges are directed and start from the responding user. Edges are annotated with the timestamp of the reply. The Slashdot dataset contains 50,824 nodes and 42,968 edges across 12 timestamps.

\subsection{Implementation details}
%Mention removing some timesteps in UCI due to very few nodes.
\subsubsection{ST-TransformerG2G}
The GCN block in Fig.~\ref{fig:ST_transformerG2G} comprises three GCN layers, each followed by a dropout layer and utilizing tanh activation. The initial GCN layer projects the input vector from an $n$-dimensional space to a $d$-dimensional vector and performs GCN updates, where $n$ represents the maximum number of nodes in each graph dataset (see the first column in Table~\ref{tab:benchmark_table} for all datasets). We use $d = 256$ for SBM, Reality Mining, Slashdot, and Bitcoin-OTC datasets and $d=512$ for the UCI dataset. For all datasets, we used a dropout rate of 0.5 in the dropout layer. In the transformer encoder, we used a single encoder layer with single-head attention. In the nonlinear layer after the encoder, a \texttt{tanh} activation function and the vectors were projected to a $512$ dimensional space, similar to our prior DynG2G work~\cite{xu2022dyng2g}. In the nonlinear projection head, we used an ELU activation function. The Adam optimizer is used to optimize the weights of the ST-TransformerG2G model across all datasets. During our experiments, we tested different lookback values (\(l = 1, 2, 3, 4, 5\)), and the optimal lookback values that achieved the best link prediction performance for each dataset are presented in Table~\ref{tab:ST_mamba_parameters}. Moreover, all relevant hyperparameters, including the learning rates (lr) for different datasets, are summarized in Table~\ref{tab:ST_mamba_parameters}.

\vspace{-.2in}
\begin{table}[H]
\centering
\caption{Hyperparameters for ST-TransformerG2G on different datasets}
\label{tab:ST_mamba_parameters}
\fontsize{11pt}{11.5pt}\selectfont 
\renewcommand{\arraystretch}{1.3} % Increase row height by 30%
\begin{tabular}{l||lllll}
\hline
Dataset & n & $d$ & lr &  $d_{out}$ & lookback \\
\hline
Reality Mining & 96 & 256 & 1e-4 & 64 & 3\\
UCI & 1899 & 512 & 1e-6 & 256 & 5 \\
SBM & 1000 & 256 & 1e-4 & 64 & 5 \\
Bitcoin-OTC & 5881 & 256 & 1e-5 & 256 & 4 \\
Slashdot & 50855 & 256 & 1e-6 & 64 & 2 \\
\hline
\end{tabular}
\end{table}

\vspace{-.1in}

The first 70\% of the timestamps was used to train the ST-TransformerG2G model, the next 10\% of the timestamps was used for validation, and 20\% for test. The validation loss was utilized to implement early stopping of the model. Once the model was trained, we used the trained model to predict and save the embeddings of the nodes for all timestamps.

\subsubsection{$\mathcal{DG}$-Mamba}
The $\mathcal{DG}$-Mamba model uses the Mamba model as its core component. The Mamba layer was configured with the hyperparameters that are listed in Table \ref{tab:DG_mamba_parameters}.
\vspace{-.2in}
\begin{table}[H]
\centering
\caption{Hyperparameters and results for $\mathcal{DG}$-Mamba on different datasets}
\label{tab:DG_mamba_parameters}
\setlength{\tabcolsep}{2pt}
\fontsize{8.5pt}{9pt}\selectfont 
\renewcommand{\arraystretch}{1.3} % Increase row height by 30%
\begin{tabular}{l||llllllll}
\hline
Dataset & $d_{model}$ & $d_{state}$ & $d_{conv}$ & lr & weight decay & intermediate space & dropout & lookback \\
\hline
Reality Mining & 76 & 6 & 9 & 0.00012 & 2.45e-05 & 64 & 0.42 & 2 \\
UCI & 1899 & 5 & 5 & 0.001 & 0.0001 & 37 & 0.2 & 4 \\
SBM & 1000 & 25 & 6 & 0.003 & 1.02e-05 & 47 & 0.32 & 2 \\
Bitcoin-OTC & 128 & 2 & 2 & 0.003 & 1.03e-03 & 15 & 0.37 & 2 \\
Slashdot & 256 & 44 & 5 & 2.76e-05 & 0.00015 & 28 & 0.44 & 2 \\
\hline
\end{tabular}
\end{table}

\vspace{-.1in}
According to Fig.~\ref{fig:TG-mamba}, the input sequence of graph snapshots is first passed through the Mamba layer to capture temporal dependencies. The output is then mean-pooled across the temporal dimension to obtain a fixed-size representation for each node. In the nonlinear layer after the Mamba block, a \texttt{tanh} activation function and the vectors were projected into a latent dimensional space, similar to the TransformerG2G paper. In the nonlinear projection head, we used an ELU activation function. To optimize the weights of the $\mathcal{DG}$-Mamba model, we used the Adam optimizer for all datasets.
The model was trained for 50 epochs using the same triplet-based contrastive loss as shown in Eq.~\ref{eq:triplet_loss}. We performed early stopping based on validation loss with a patience of 10 epochs. For a fair comparison, we used the same embedding size (i.e., $d_{out}$ in Table~\ref{tab:ST_mamba_parameters}) for each dataset as was employed during the training of the ST-TransformerG2G model. The first 70\% of the timestamps were used for training the $\mathcal{DG}$-Mamba model, and the next 10\% of the timestamps were used for validation. The validation loss was used to perform early
stopping of the model. Once the model was trained, we used the trained model to predict and save the node embeddings for all timestamps.

\subsubsection{$\mathcal{GDG}$-Mamba}

The $\mathcal{GDG}$-Mamba model incorporates Graph Isomorphism Network Edge (GINE) convolutions to enhance spatial representation by incorporating crucial edge features before processing temporal dynamics with the Mamba layer. The hyperparameters used to configure the $\mathcal{GDG}$-Mamba model during its implementation are presented in Table \ref{tab:GDG_mamba_params}.

\vspace{-.2in}
\begin{table}[h]
\centering
\caption{Hyperparameters for the $\mathcal{GDG}$-Mamba model with GINE Conv and Mamba}
\label{tab:GDG_mamba_params}
\setlength{\tabcolsep}{2pt}
\fontsize{8.5pt}{9pt}\selectfont 
\renewcommand{\arraystretch}{1.3} % Increase row height by 30%
\begin{tabular}{l||cccccccc}
\hline
Dataset & $d_{model}$ & $d_{state}$ & $d_{conv}$ & lr & weight decay & intermediate space & dropout & lookback \\
\hline
Reality Mining & 76 & 6 & 3 & 2.20e-05 & 1.46e-05 & 49 & 0.176 & 2 \\
UCI & 1899 & 3 & 7 & 0.00014 & 0.00025 & 8 & 0.41 & 3 \\
SBM & 1000 & 18 & 4 & 0.000139 & 0.0005 & 15 & 0.18 & 1 \\
Bitcoin-OTC & 128 & 45 & 3 & 3.65e-05 & 3.98e-05 & 110 & 0.41 & 3 \\
Slashdot & 256 & 2 & 5 & 2.76e-05 & 0.00015 & 28 & 0.44 & 4 \\
\hline
\end{tabular}
\end{table}

\vspace{-.1in}
The model was trained over 50 epochs using a triplet-based contrastive loss and the Adam optimizer, incorporating early stopping based on validation loss to optimize performance. We applied the same amount of training data (70\%), evaluation data (10\%) for the evaluation of ST-transformerG2G, $\mathcal{DG}$-Mamba, and $\mathcal{GDG}$-Mamba. The parameters presented in Tables~\ref{tab:DG_mamba_parameters} and \ref{tab:GDG_mamba_params} were determined through a comprehensive fine-tuning process. To identify the optimal hyperparameters for our models, we utilized Optuna~\cite{optuna_2019}, a hyperparameter optimization framework. We defined a specific search space for each hyperparameter and employed Optuna's Tree-structured Parzen Estimator (TPE) sampler to navigate this space effectively. Our primary goal was to maximize the Mean Average Precision (MAP) for the temporal link prediction task, a critical metric for assessing the performance of dynamic graph embedding models.

\subsection{Temporal link prediction results}

To evaluate node embeddings for link prediction, we employ a logistic regression classifier that uses concatenated node embeddings to estimate the probability of a link between them. 
The classifier is trained on node pairs labeled as linked or unlinked, sampled as positive (existing edges) and negative (non-existing edges) from each graph snapshot. To balance the training data, we maintain a ratio of 1:10 for positive to negative samples.

Mean Average Precision (MAP) loss was used for link prediction, which improves link prediction by prioritizing the ranking of true links over non-links. The classifier is trained with the Adam optimizer and a learning rate of 1e-3. Moreover, the performance of the link prediction classifier is evaluated using two key metrics: MAP and Mean Reciprocal Rank (MRR). MAP measures the average precision across all queries, providing an overall measure of link prediction accuracy. MRR calculates the average of the reciprocal ranks of the first relevant item for each query, focusing on the model's ability to rank true links at the top. The corresponding link prediction results for Reality Mining, UCI, SBM, Bitcoin-OTC and Slashdot datasets, are presented in the Tables \ref{tab:realitymining_results}, \ref{tab:uci_results}, \ref{tab:sbm_results}, \ref{tab:bitcoin_results}, and \ref{tab:slashdot_results}, respectively. All results are based on five random initializations. 

\textbf{Reality Mining} (Table~\ref{tab:realitymining_results}): $\mathcal{GDG}$-Mamba consistently outperforms all other models. This likely stems from its ability to effectively leverage both node and edge features through GINEConv, which might be crucial for capturing the nuances of human contact patterns.

\textbf{UCI} (Table~\ref{tab:uci_results}): $\mathcal{GDG}$-Mamba again demonstrates superior performance. Its success can likely be attributed to the same reasons as in Reality Mining, highlighting its ability to handle highly transient dynamics in online social interactions where edge features play a significant role.

\textbf{SBM} (Table~\ref{tab:sbm_results}):  Both $\mathcal{DG}$-Mamba and $\mathcal{GDG}$-Mamba achieve comparable performance to ST-TransformerG2G, showing that Mamba-based models can effectively capture temporal patterns even in synthetic datasets. This suggests that the inductive biases of state-space models are well-suited for learning the evolving community structures in SBM.

\textbf{Bitcoin-OTC} (Table~\ref{tab:bitcoin_results}): $\mathcal{GDG}$-Mamba shows the best performance, likely because it effectively captures the complex transaction patterns and evolving trust relationships in the Bitcoin network by incorporating edge features.

\textbf{Slashdot} (Table~\ref{tab:slashdot_results}): ST-TransformerG2G exhibits the best performance, indicating the benefits of incorporating GCNs for learning spatial dependencies in social networks. However, $\mathcal{GDG}$-Mamba shows lower performance, potentially due to overfitting to the limited number of timestamps and the sparsity of the dataset. The edge features in Slashdot might also be less informative for the link prediction task, making the added complexity of GINEConv less beneficial.
% \vspace{-25in}
\vspace{-.4cm}
\begin{table}[H]
\centering
\caption{Link prediction performance for Reality Mining in terms of MAP and MRR for different lookbacks ($l$).}
\vspace{.05cm}
\label{tab:realitymining_results}
\scalebox{0.59}{
\setlength{\tabcolsep}{3pt}
\renewcommand{\arraystretch}{1.3} % Increase row height by 30%
\begin{tabular}{ c || c c c c c c }
\hline
Model & Metric  & $l = 1$  & $l = 2$ & $l = 3$   & $l = 4$ & $l = 5$ \\
\hline
\hline
% SBM
\multicolumn{1}{ c ||}{\multirow{2}{*}{TransformerG2G \cite{varghese2024transformerg2g}}} & MAP & 0.2010 ± 0.0460 & 0.2057 ± 0.0356 &  0.2153 ± 0.0198 & \textbf{0.2252 ± 0.0137} & 0.1965 ± 0.0349\\ 
\multicolumn{1}{ c ||}{}                     & MRR & 0.1414 ± 0.0046 & 0.1306 ± 0.0094 & 0.1334 ± 0.0094 & 0.1294 ± 0.0129 & 0.1311 ± 0.0083\\\hline
\multicolumn{1}{ c ||}{\multirow{2}{*}{ST-TransformerG2G}} & MAP & 0.2034 ± 0.0182 & 0.1994 ± 0.0334 & \textbf{0.2109 ± 0.0254} & 0.1996 ± 0.0253 & 0.1927 ± 0.0284\\ 
\multicolumn{1}{ c ||}{}                     & MRR & 0.2581 ± 0.0115 & 0.2281 ± 0.0449 & 0.2382 ± 0.0366 & 0.2433 ± 0.0152 & 0.2298 ± 0.0383\\\hline
\multicolumn{1}{ c ||}{\multirow{2}{*}{$\mathcal{DG}$-Mamba}} & MAP & 0.2997 ± 0.0023 & 0.3170 ± 0.0019 & \textbf{0.3268 ± 0.0012} & 0.2949 ± 0.0029 & 0.3076 ± 0.0016\\ 
\multicolumn{1}{ c ||}{}                     & MRR & 0.0964 ± 0.000046 & 0.1606 ± 0.002 & 0.1669 ± 0.0013 & 0.1580 ± 0.0026 & 0.1737 ± 0.0006\\\hline
\multicolumn{1}{ c ||}{\multirow{2}{*}{$\mathcal{GDG}$-Mamba}} & MAP & 0.4728 ± 0.0022 & \textbf{0.4968 ± 0.0017} & 0.4869 ± 0.0023 & 0.4808 ± 0.0018 & 0.4620 ± 0.0087\\ 
\multicolumn{1}{ c ||}{}                     & MRR & 0.2834 ± 0.0042 & 0.3057 ± 0.0024 & 0.2966 ± 0.0014 & 0.2979 ± 0.0009 & 0.2827 ± 0.0002\\\hline
\end{tabular}}
\end{table}

\vspace{-.6cm}
%%%%%%%%%%%%%%%%%%%%%%%%%%%%%%%
%%%%%%%%%%%%% UCI %%%%%%%%%%%%%
%%%%%%%%%%%%%%%%%%%%%%%%%%%%%%%
\begin{table}[H]
\centering
\caption{Link prediction performance for UCI in terms of MAP and MRR for different lookbacks ($l$).}
\vspace{.05cm}
\label{tab:uci_results}
\scalebox{0.59}{
\setlength{\tabcolsep}{3pt}
\renewcommand{\arraystretch}{1.3} % Increase row height by 30%
\begin{tabular}{ c || c c c c c c }
\hline
Model & Metric  & $l = 1$  & $l = 2$ & $l = 3$   & $l = 4$ & $l = 5$ \\
\hline
\hline
% SBM
\multicolumn{1}{ c ||}{\multirow{2}{*}{TransformerG2G\cite{varghese2024transformerg2g}}} & MAP & 0.0241 ± 0.0152 & 0.0347± 0.0222 & 0.0340 ± 0.0256 & 0.0342 ± 0.0244 & \textbf{0.0495 ± 0.0107}\\
\multicolumn{1}{ c ||}{}                     & MRR & 0.2616 ± 0.0879 & 0.2947 ± 0.1087 & 0.2905 ± 0.1121 & 0.3042 ± 0.1115 & 0.3447 ± 0.0196\\\hline
\multicolumn{1}{ c ||}{\multirow{2}{*}{ST-TransformerG2G}} & MAP & 0.0392 ± 0.0252 & 0.0417 ± 0.0350 & 0.0390 ± 0.0214 & 0.0472 ± 0.0264 & \textbf{0.0604 ± 0.0299}\\
\multicolumn{1}{ c ||}{}                     & MRR & 0.2403 ± 0.0747 & 0.2789 ± 0.1019 & 0.2929 ± 0.0392 & 0.3211 ± 0.0398 & 0.3322 ± 0.0382\\\hline
\multicolumn{1}{ c ||}{\multirow{2}{*}{$\mathcal{DG}$-Mamba}} & MAP & 0.0766 ± 0.0034 & \textbf{0.0798 ± 0.0013} & 0.0652 ± 0.0015 & 0.1113 ± 0.0019 & 0.0679 ± 0.002\\ 
\multicolumn{1}{ c ||}{}                     & MRR & 0.3772 ± 0.0015 & 0.4470 ± 0.0031 & 0.4250 ± 0.00192 & 0.4670 ± 0.0038 & 0.4504 ± 0.0022\\\hline
\multicolumn{1}{ c ||}{\multirow{2}{*}{$\mathcal{GDG}$-Mamba}} & MAP & 0.1872 ± 0.0032 & 0.2108 ± 0.0023 & \textbf{0.2266 ± 0.0023} & 0.1791 ± 0.0035 & 0.1537 ± 0.0061\\ 
\multicolumn{1}{ c ||}{}                     & MRR & 0.5143 ± 0.002 & 0.5074 ± 0.0024 & 0.5339 ± 0.0034 & 0.5241 ± 0.0029 & 0.5163 ± 0.0009\\\hline
\end{tabular}}
\end{table}

\vspace{-.6cm}
%%%%%%%%%%%%%%%%%%%%%%%%%%%%%%%
%%%%%%%%%%%%% SBM %%%%%%%%%%%%%
%%%%%%%%%%%%%%%%%%%%%%%%%%%%%%%
\begin{table}[H]
\centering
\caption{Link prediction performance for SBM in terms of MAP and MRR for different lookbacks ($l$).}
\vspace{.05cm}
\label{tab:sbm_results}
\scalebox{0.59}{
\setlength{\tabcolsep}{3pt}
\renewcommand{\arraystretch}{1.3} % Increase row height by 30%
\begin{tabular}{ c || c c c c c c }
\hline
Model & Metric  & $l = 1$  & $l = 2$ & $l = 3$   & $l = 4$ & $l = 5$ \\
\hline
\hline
% SBM
\multicolumn{1}{ c ||}{\multirow{2}{*}{TransformerG2G\cite{varghese2024transformerg2g}}} & MAP & \textbf{0.6204 $\pm$ 0.0386} & 0.6143 ± 0.0274 &  0.5927 ± 0.0192 & 0.6096 ± 0.0360 & 0.6097 ± 0.0104\\ 
\multicolumn{1}{ c ||}{}                     & MRR & 0.0369 $\pm$ 0.0055 & 0.0388 ± 0.0009 &0.0317 ± 0.0053 & 0.0349 ± 0.0059 & 0.0386 ± 0.0002\\\hline
\multicolumn{1}{ c ||}{\multirow{2}{*}{ST-TransformerG2G}} & MAP & 0.6717 ± 0.0246 & 0.6425 ± 0.0196 & \textbf{0.6721 ± 0.0225} & 0.6355 ± 0.0110 & 0.6577 ± 0.0259\\ 
\multicolumn{1}{ c ||}{}                     & MRR & 0.0388 ± 0.0050 & 0.0328 ± 0.0051 & 0.0391 ± 0.0048 & 0.0326 ± 0.0042 & 0.0371 ± 0.0061\\\hline
\multicolumn{1}{ c ||}{\multirow{2}{*}{$\mathcal{DG}$-Mamba}} & MAP & 0.6896 ± 0.0013 & \textbf{0.6898 ± 0.0012} & 0.6748 ± 0.0005 & 0.6720 ± 0.0001 & 0.6773 ± 0.0019\\ 
\multicolumn{1}{ c ||}{}                     & MRR & 0.0418 ± 0.0001 & 0.0424 ± 0.0001 & 0.0421 ± 0.0001 & 0.0412 ± 0.0001 & 0.0408 ± 0.0002\\\hline
\multicolumn{1}{ c ||}{\multirow{2}{*}{$\mathcal{GDG}$-Mamba}} & MAP & \textbf{0.6928 ± 0.0042} & 0.6828 ± 0.0045 & 0.6863 ± 0.0013 & 0.6872 ± 0.0035 & 0.6857 ± 0.0061\\ 
\multicolumn{1}{ c ||}{}                     & MRR & 0.0420 ± 0.0020 & 0.0413 ± 0.0026 & 0.0415 ± 0.0031 & 0.0422 ± 0.0019 & 0.0416 ± 0.0001\\\hline
\end{tabular}}
\end{table}

\vspace{-.6cm}
%%%%%%%%%%%%%%%%%%%%%%%%%%%%%%%
%%%%%%%%%%%%% Bitcoin %%%%%%%%%%%%%
%%%%%%%%%%%%%%%%%%%%%%%%%%%%%%%
\begin{table}[H]
\centering
\caption{Link prediction performance for Bitcoin-OTC in terms of MAP and MRR for different lookbacks ($l$).}
\vspace{.05cm}
\label{tab:bitcoin_results}
\scalebox{0.59}{
\setlength{\tabcolsep}{3pt}
\renewcommand{\arraystretch}{1.3} % Increase row height by 30%
\begin{tabular}{ c || c c c c c c }
\hline
Model & Metric  & $l = 1$  & $l = 2$ & $l = 3$   & $l = 4$ & $l = 5$ \\
\hline
\hline
% SBM
\multicolumn{1}{ c ||}{\multirow{2}{*}{TransformerG2G\cite{varghese2024transformerg2g}}} & MAP & 0.0278 ± 0.0034 & 0.0274 ± 0.0032 & 0.0274 ± 0.0032 & \textbf{0.0319 ± 0.0066} & 0.0303 ± 0.0042\\ 
\multicolumn{1}{ c ||}{}                     & MRR & 0.3194 ± 0.0182 & 0.3415 ± 0.0246 & 0.3415 ± 0.0246 & 0.4089 ± 0.0079 & 0.4037 ± 0.0188\\\hline
\multicolumn{1}{ c ||}{\multirow{2}{*}{ST-TransformerG2G}} & MAP & 0.0498 ± 0.0283 & 0.0446 ± 0.0291 & 0.0665 ± 0.0191 & \textbf{0.0787 ± 0.0093} & 0.0409 ± 0.0320\\ 
\multicolumn{1}{ c ||}{}                     & MRR & 0.3435 ± 0.1027 & 0.3321 ± 0.0620 & 0.4048 ± 0.0482 & 0.4252 ± 0.0474 & 0.3357 ± 0.0996\\\hline
\multicolumn{1}{ c ||}{\multirow{2}{*}{$\mathcal{DG}$-Mamba}} & MAP & 0.0759 ± 0.0011 & \textbf{0.1331 ± 0.00001} & 0.0566 ± 0.0003 & 0.0602 ± 0.0023 & 0.05012 ± 0.0005\\ 
\multicolumn{1}{ c ||}{}                     & MRR & 0.4404 ± 0.0018 & 0.5369 ± 0.00002 & 0.5063 ± 0.0002 & 0.5085 ± 0.0014 & 0.5526 ± 0.0005\\\hline
\multicolumn{1}{ c ||}{\multirow{2}{*}{$\mathcal{GDG}$-Mamba}} & MAP & 0.0827 ± 0.0009 & 0.0560 ± 0.0013 & \textbf{0.1822 ± 0.0003} & 0.0919 ± 0.0005 & 0.1522 ± 0.0007\\ 
\multicolumn{1}{ c ||}{}                     & MRR & 0.3837 ± 0.0012 & 0.4666 ± 0.0004 & 0.5367 ± 0.0009 & 0.531 ± 0.0006 & 0.5795 ± 0.0011\\\hline
\end{tabular}}
\end{table}

\vspace{-.6cm}
%%%%%%%%%%%%%%%%%%%%%%%%%%%%%%%
%%%%%%%%%%%%% Slashdot %%%%%%%%%%%%%
%%%%%%%%%%%%%%%%%%%%%%%%%%%%%%%
\begin{table}[H]
\centering
\caption{Link prediction performance for Slashdot in terms of MAP and MRR for different lookbacks ($l$).}
\vspace{.05cm}
\label{tab:slashdot_results}
\scalebox{0.59}{
\setlength{\tabcolsep}{3pt}
\renewcommand{\arraystretch}{1.3} % Increase row height by 30%
\begin{tabular}{ c || c c c c c c }
\hline
Model & Metric  & $l = 1$  & $l = 2$ & $l = 3$   & $l = 4$ & $l = 5$ \\
\hline
\hline
% SBM
\multicolumn{1}{ c ||}{\multirow{2}{*}{TransformerG2G\cite{varghese2024transformerg2g}}} & MAP & \textbf{0.0498 ± 0.0059} & 0.0343 ± 0.0147 & 0.0360 ± 0.0137 & 0.0368 ± 0.0183 & 0.0268 ± 0.0153\\ 
\multicolumn{1}{ c ||}{}                     & MRR & 0.2568 ± 0.0126 & 0.2679 ± 0.0222 & 0.2824 ± 0.0143 & 0.2783 ± 0.0358 & 0.2749 ± 0.0096\\\hline
\multicolumn{1}{ c ||}{\multirow{2}{*}{ST-TransformerG2G}} & MAP & \textbf{0.1081 ± 0.0613} & 0.0860 ± 0.0444 & 0.0572 ± 0.0437 & 0.0499 ± 0.0280 & 0.0432 ± 0.0352\\ 
\multicolumn{1}{ c ||}{}                     & MRR & 0.3979 ± 0.0582 & 0.3968 ± 0.0379 & 0.3711 ± 0.0719 & 0.3656 ± 0.0599 & 0.3429 ± 0.0928\\\hline
\multicolumn{1}{ c ||}{\multirow{2}{*}{$\mathcal{DG}$-Mamba}} & MAP & 0.0576 ± 0.0013 & 0.0651 ± 0.0012 & \textbf{0.0676 ± 0.0009} & 0.0555 ± 0.0022 & 0.0006 ± 0.0003\\ 
\multicolumn{1}{ c ||}{}                     & MRR & 0.3310 ± 0.0001 & 0.3675 ± 0.0001 & 0.3814 ± 0.0006 & 0.3529 ± 0.00008 & 0.1082 ± 0.0001\\\hline
\multicolumn{1}{ c ||}{\multirow{2}{*}{$\mathcal{GDG}$-Mamba}} & MAP & 0.0216 ± 0.0012 & \textbf{0.03992 ± 0.0011} & 0.0245 ± 0.0032 & 0.0331 ± 0.0052 & 0.0129 ± 0.0021\\ 
\multicolumn{1}{ c ||}{}                     & MRR & 0.2285 ± 0.0032 & 0.2052 ± 0.0032 & 0.2529 ± 0.0011 & 0.2353 ± 0.0012 & 0.2397 ± 0.0007\\\hline
\end{tabular}}
\end{table}

\begin{figure}[H]
    \centering
    \includegraphics[scale=0.7]{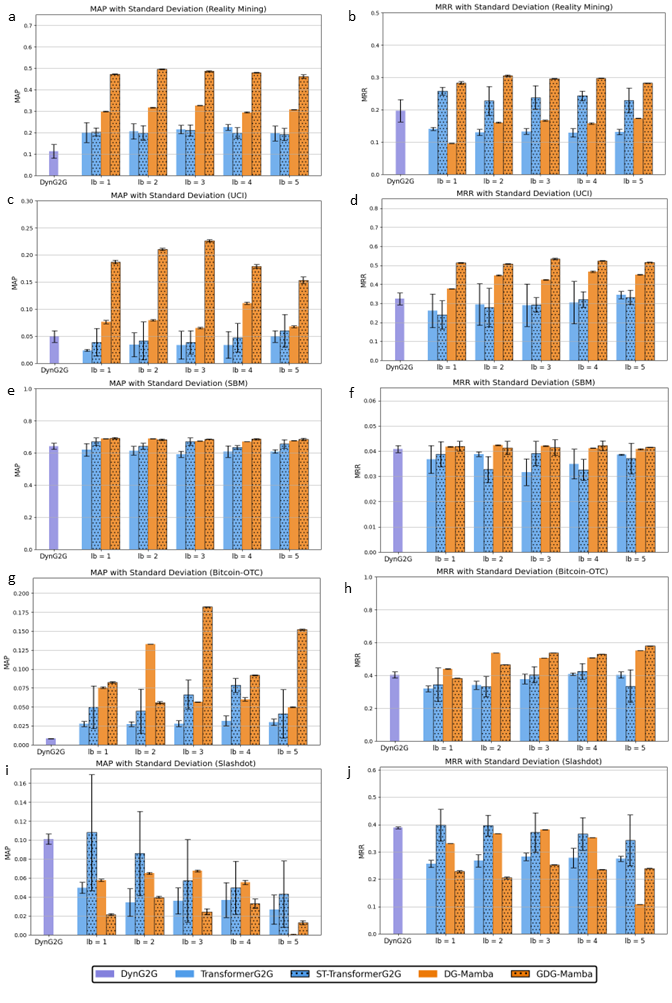}
    \vspace{-0.75cm}
    \caption{Comparison of MAP (1st column) and MRR (2nd colum) for temporal link prediction task for DynG2G, TransformerG2G, ST-TransformerG2G, $\mathcal{DG}$-Mamba, $\mathcal{GDG}$-Mamba models.} 
    \label{fig:map_mrr_figs}
\end{figure}

\vspace{-.2cm}
The bar plot figure below offers a detailed visualization of each model's best performance in the temporal link prediction task, comparing the capabilities of our prior two models—{\it DynG2G and TransformerG2G}—with three new proposed models: {\it ST-TransformerG2G, $\mathcal{DG}$-Mamba, and $\mathcal{GDG}$-Mamba}. This comparison highlights the respective strengths of each model in handling dynamic graph data. For a comprehensive view of the results, including error bars that illustrate variability based on five random initializations, see Figure \ref{fig:map_mrr_figs}. 

Note that the $\mathcal{DG}$-Mamba and $\mathcal{GDG}$-Mamba models demonstrated superior and more stable (i.e., smaller standard deviations) performance in link prediction across most benchmarks, except for the Slashdot dataset, when compared to the ST-TransformerG2G and TransformerG2G models. The underperformance of $\mathcal{GDG}$-Mamba with GINEConv on the Slashdot dataset could be due to a combination of factors related to the dataset's unique characteristics and the model's complexity. The limited number of timestamps in Slashdot (only 12) might make $\mathcal{GDG}$-Mamba, which incorporates the additional complexity of GINEConv for integrating edge features, prone to overfitting. Moreover, the sparsity and specific temporal dynamics of interactions within the Slashdot dataset could challenge $\mathcal{GDG}$-Mamba's ability to effectively capture the relevant information necessary for accurate link prediction. Moreover, the optimal lookback value that achieves the best performance varies for each benchmark, reflecting the model's ability to capture the inherent temporal dynamics specific to each dataset.

\begin{table}[!h]
\caption{Comparison results of the temporal link prediction task for five different benchmarks.}
\vspace{0.1in}
\label{tab:method_comparisons_LP}
\centering
\setlength{\tabcolsep}{2pt}
\fontsize{8pt}{9pt}\selectfont 
\begin{tabular}{llll}
\hline
Benchmark              & Method                     & MAP             & MRR    \\
\hline \hline

\multirow{2}{*}{Reality Mining}   

                                & EvolveGCN                  & 0.0090          & 0.0416 \\
                                & DynG2G  &  0.1126 ± 0.0322  & 0.1973 ± 0.0341                   \\
                                & TransformerG2G ($l = 4$)   & 0.2252 ± 0.0137  & 0.1294 ± 0.0129\\
                                & ST-TransformerG2G ($l=3$) & 0.2109 ± 0.0254 & 0.2382 ± 0.0366 \\
                                & $\mathcal{DG}$-Mamba ($l=3$)             & 0.3268 ± 0.0012 & 0.1669 ± 0.0013\\
                                & $\mathcal{GDG}$-Mamba ($l = 2$) & \textbf{0.4968 ± 0.0017} & \textbf{0.3057 ± 0.0024}\\
\hline
\multirow{5}{*}{UCI}         & DynGEM                     & 0.0209          & 0.1055                     \\
                             & dyngraph2vecAE             & 0.0044          & 0.0540                     \\
                             & dyngraph2vecAERNN          & 0.0205          & 0.0713                     \\
                             & EvolveGCN                  & 0.0270          & 0.1379                     \\
                             & DynG2G                    & $0.0348\pm 0.0121$          & $0.3247\pm 0.0312$                    \\
                             & TransformerG2G ($l = 5$)   & 0.0495 ± 0.0107         & 0.3447 ± 0.0196\\
                             & ST-TransformerG2G ($l=5$) & 0.0604 ± 0.0299  & 0.3322 ± 0.0382\\
                             & $\mathcal{DG}$-Mamba ($l=2$)             & 0.0798 ± 0.0013  & 0.4470 ± 0.0031\\
                             & $\mathcal{GDG}$-Mamba ($l =3$) & \textbf{0.2266 ± 0.0023} & \textbf{0.5339 ± 0.0034}\\
\hline
\multirow{5}{*}{SBM} & DynGEM                     & 0.1680           & 0.0139 \\
                     & dyngraph2vecAE             & 0.0983          & 0.0079 \\
                     & dyngraph2vecAERNN          & 0.1593          & 0.0120  \\
                     & EvolveGCN                  & 0.1989          & 0.0138 \\
                     & DynG2G           &  0.6433 $\pm$ 0.0187   & 0.0409 $\pm$ 0.0014\\
                     & TransformerG2G ($l = 1$)   & $0.6204 \pm 0.0386$        & 0.0369 $\pm$ 0.0055\\
                     & ST-TransformerG2G ($l=3$) & 0.6721 ± 0.0225 & 0.0391 ± 0.0048\\
                     & $\mathcal{DG}$-Mamba ($l=2$)             &0.6898 ± 0.0012& \textbf{0.0424 ± 0.0001} \\
                     & $\mathcal{GDG}$-Mamba ($l =1$) & \textbf{0.6928 ± 0.0042} & 0.0420 ± 0.0020\\
\hline
\multirow{5}{*}{Bitcoin-OTC} & DynGEM                     & 0.0134          & 0.0921                     \\
                             & dyngraph2vecAE             & 0.0090          & 0.0916 \\
                             & dyngraph2vecAERNN & 0.0220 & 0.1268                     \\
                              & EvolveGCN                  & 0.0028          & 0.0968                     \\
                             & DynG2G                   & 0.0083 ± 0.0027       &0.3529 ± 0.0100                         \\
                             & TransformerG2G ($l = 4$)   & 0.0303 ± 0.0042         &  0.4037 ± 0.0188\\
                             & ST-TransformerG2G ($l=$4) &  0.0787 ± 0.0093 & 0.4252 ± 0.0474 \\
                             & $\mathcal{DG}$-Mamba ($l=2$)             & 0.1331 ± 0.00001 & \textbf{0.5369 ± 0.00002}\\
                             & $\mathcal{GDG}$-Mamba ($l =3$) & \textbf{0.1822 ± 0.0003} & 0.5367 ± 0.0009\\
\hline
\multirow{1}{*}{Slashdot}  & DynG2G                  & 0.1012 ± 0.0053         & 0.3885 ± 0.0030                     \\
& TransformerG2G ($l = 1$)   & 0.0498 ± 0.0059         & 0.2568 ± 0.0126\\
                     & ST-TransformerG2G ($l=1$) & \textbf{0.1081 ± 0.0613}  & \textbf{0.3979 ± 0.0582} \\
                     & $\mathcal{DG}$-Mamba ($l=3$)             & 0.0676 ± 0.0009 & 0.3814 ± 0.0006\\
                     & $\mathcal{GDG}$-Mamba ($l = 2$) & 0.0399 ± 0.0011 & 0.2052 ± 0.0032\\
\hline \hline
\end{tabular}
\end{table}

%\vspace{-.1in}
Furthermore, we compare the performance of our three proposed models with six baseline models in link prediction tasks. As shown in Table~\ref{tab:method_comparisons_LP}, the Mamba-based models ($\mathcal{DG}$-Mamba and $\mathcal{GDG}$-Mamba) consistently outperformed transformer-based approaches and other baselines in temporal link prediction tasks, particularly for the Bitcoin-OTC, UCI, and Reality Mining datasets, with highly transient dynamics. However, due to the limited number of timesteps in the Slashdot dataset, the Mamba-based models achieved lower performance compared to the transformer-based models. With the SBM dataset, our proposed models achieved similar performance, as SBM represents a stable temporal graph.
\begin{itemize}
    \item \textit{DynGEM~\cite{goyal2018dyngem}}: This is a dynamic graph embedding method that uses autoencoders to obtain graph embeddings. The weights of the previous timestamp are used to initialize the weights of the current timestamp, and thereby make the training process faster.
    \item \textit{dyngraph2vecAE~\cite{goyal2020dyngraph2vec}}: This method uses temporal information in dynamic graphs to obtain embeddings using an autoencoder. The encoder takes in the historical information of the graph, and the decoder reconstructs the graph at the next timestamp, and the latent vector corresponds to the embedding of the current timestamp.
    \item \textit{dyngraph2vecAERNN~\cite{goyal2020dyngraph2vec}}: This method is similar to dyngraph2vecAE, except that the feed-forward neural network in the encoder is replaced by LSTM layers. This method also incorporates historical information while predicting graph embeddings.
    \item \textit{EvolveGCN~\cite{pareja2020evolvegcn}}: A graph convolution network (GCN) forms the core of this model. A recurrent neural network (RNN) is used to evolve the GCN parameters along time.
    \item \textit{ROLAND~\cite{you2022roland}}: A graph neural network is combined with GRU to learn the temporal dynamics. The GNN has attention layers to capture the spatial-level node importance.
    \item \textit{DynG2G~\cite{xu2022dyng2g}}: This method uses the G2G encoder at its core to obtain the embeddings. The weights from the previous timestamp are used to initialize the weights of the next timestamp to make the training faster. Unlike the previous methods, DynG2G includes information about the uncertainty of the embeddings.
\end{itemize}

\subsection{Analysis of the State Matrix A in the Mamba-based Models}

To gain deeper understanding into how the Mamba model captures temporal dependencies in dynamic graphs, we analyze the learned state transition matrix $A$, which plays a crucial role in the state-space model formulation and in capturing temporal dynamics. In our prior work~\cite{varghese2024transformerg2g}, time-dependent attention matrices have been explored for the adaptive stepping. The hidden attention mechanism underlying the Mamba model was proposed to reformulate the computation in the Mamba layer, revealing implicit attention-like mechanisms~\cite{hidden_mamba}.

\begin{equation}
\label{eq:mamba_attention}
\tilde{\alpha}_{i,j} \approx \tilde{Q}_i \tilde{H}_{i,j} \tilde{K}_j,
\end{equation}
where $\tilde{Q}_i = S_C(\hat{x}_i), 
\tilde{K}_j = \sigma(S_\Delta(\hat{x}_j)) S_B(\hat{x}_j)$, and
\begin{align*}
\tilde{H}_{i,j} &= \exp\left(
    \sum_{\substack{k = j+1 \\ S_\Delta(\hat{x}_k) > 0}}^i 
    S_\Delta(\hat{x}_k)
\right) A.
\end{align*}

Here, $\tilde{\alpha}_{i,j}$ represents the influence of the $j$-th timestamp on the $i$-th timestamp, analogous to attention weights in the transformer models. $\tilde{Q}_i$ and $\tilde{K}_j$ can be interpreted as query and key vectors, while $\tilde{H}_{i,j}$ encodes the cumulative historical context between timestamps $j$ and $i$. 

\begin{figure}[H]
    \centering
    \includegraphics[width =.98\textwidth, height=8.5cm]{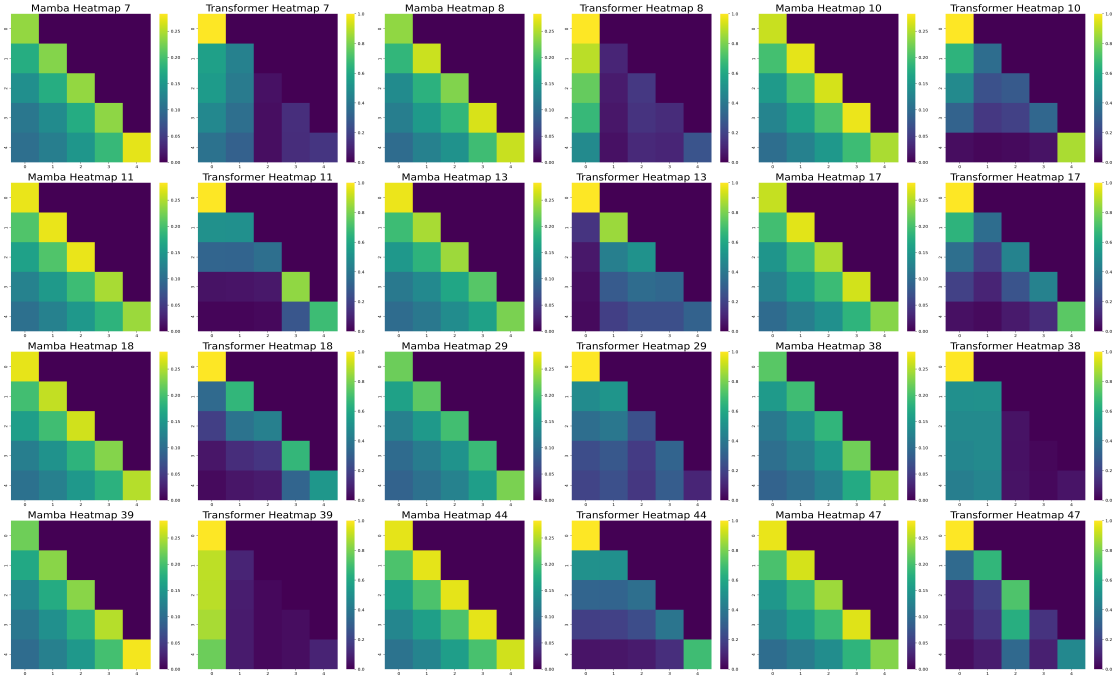} 
    \caption{Comparison of $\mathcal{DG}$-Mamba state transition matrices (Columns 1, 3, 5) and TransformerG2G attention matrices (Columns 2, 4, 6) across 24 randomly selected time steps for Reality Mining. }
    \label{fig:statematrixA}
\end{figure}

\begin{figure}[H]
    \centering
    \includegraphics[width=\textwidth,height = 10cm]{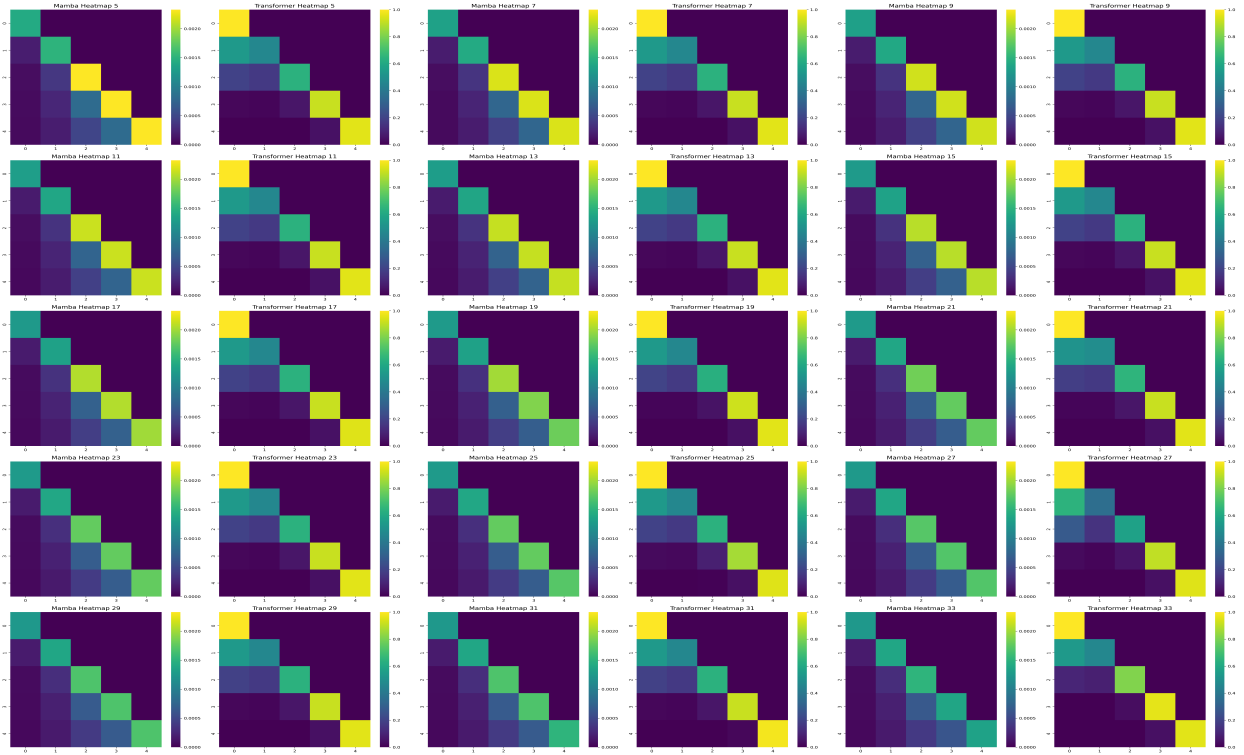} 
    \caption{Comparison of $\mathcal{DG}$-Mamba state transition matrix and TransformerG2G attention matrix for SBM for odd timestamps from 5 to 33.}
    \label{fig:enter-label}
\end{figure}

Figures \ref{fig:statematrixA} and \ref{fig:enter-label} compare the $\mathcal{DG}$-Mamba state transition matrix (left column) and the TransformerG2G attention matrix (right column) for the Reality Mining and SBM datasets, respectively, using 24 random timestamps for Reality Mining and all odd timestamps from 5 to 33 for SBM. Each row represents a timestamp, and each column represents the historical timestamps considered (lookback). Warmer colors indicate stronger dependencies. Below are the key observations:

\begin{itemize}
    \item {\it $\mathcal{DG}$-Mamba demonstrates a strong ability to capture long-term temporal information}, as evidenced by the significant dependencies extending beyond recent timestamps. This is particularly noticeable in the Reality Mining dataset, where connections change rapidly.
    \item {\it While TransformerG2G also captures long-term dependencies}, its pattern is more diffuse, indicating a broader but less focused consideration of historical information. 
    \item {\it The differences are more pronounced in the Reality Mining dataset}, which exhibits more dynamic temporal behavior compared to the relatively stable SBM dataset. In SBM, both models show a greater emphasis on recent timestamps, although $\mathcal{DG}$-Mamba still maintains stronger long-range dependencies.
\end{itemize}

These observations highlight the contrasting approaches to temporal modeling: $\mathcal{DG}$-Mamba effectively captures long-term dependencies while maintaining efficiency, making it particularly well-suited for dynamic graphs with rapid changes like Reality Mining. TransformerG2G, while capable of capturing long-range context, may be less focused and potentially incur higher computational costs. The choice of model depends on the specific dataset characteristics as well as the desired trade-off between capturing long-term dependencies and computational efficiency.

\subsection{Computational Efficiency Comparisons and Analysis}
\begin{figure}[H]
    \centering    
    \includegraphics[width=.45\textwidth]{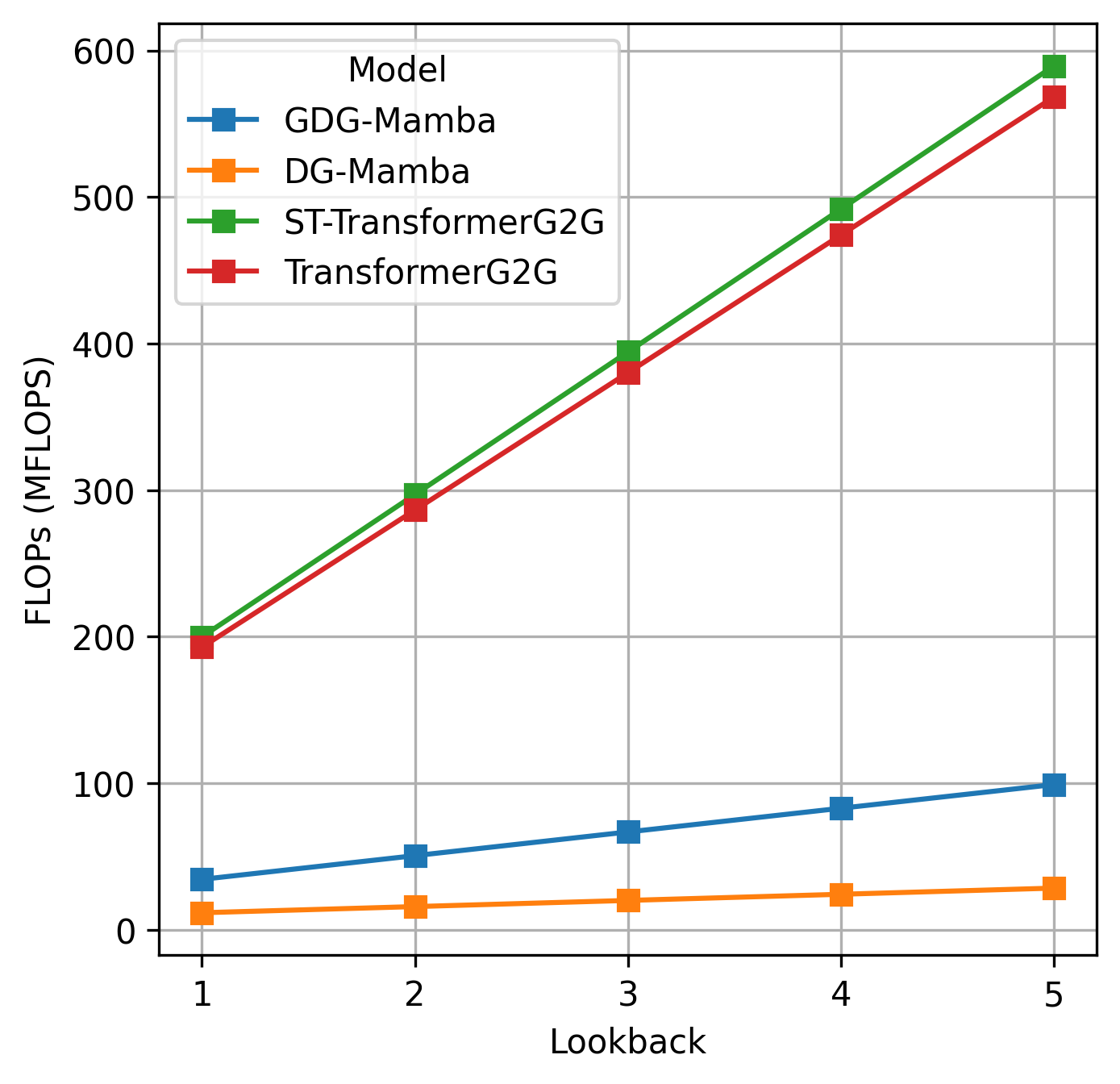} \quad \includegraphics[width=.45\textwidth]{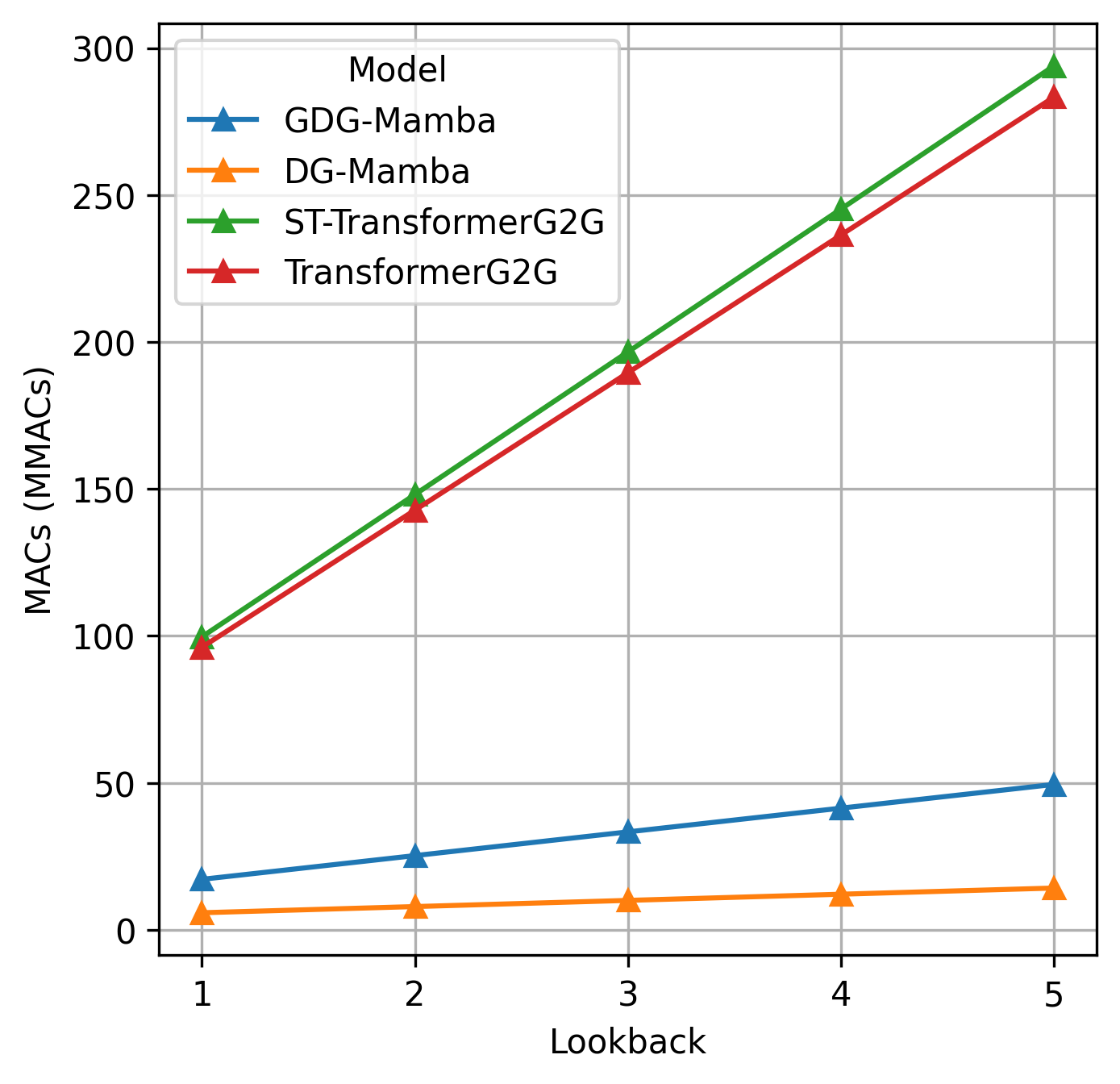} 
    \caption{Computational cost comparison of $\mathcal{GDG}$-Mamba, $\mathcal{DG}$-Mamba, TransformerG2G, and $\mathcal{ST}$-TransformerG2G as a function of the lookback window size. Left and Right display the measured FLOPs and MACs values over lookbacks.}
    \label{fig:flops-macs}
\end{figure}

We compare the computational cost of $\mathcal{GDG}$ -Mamba, $\mathcal{DG}$ -Mamba, TransformerG2G, and $\mathcal{ST}$ -TransformerG2G with increasing lookback windows. As shown in Figures~\ref{fig:flops-macs} (a) and (b), both FLOPs and MACs scale linearly for Mamba-based models, while Transformer-based baselines exhibit quadratic growth. Notably, $\mathcal{ST}$-TransformerG2G is marginally more expensive than TransformerG2G at each lookback. These results empirically validate the efficiency of state-space models for temporal graph encoding under long-range dependencies.

\section{Conclusion} 
\label{conclusions}
This study provides a comparative analysis of dynamic graph embedding leveraging approaches using state-space models and transformers, focusing on their scalability, efficiency, and predictive performance across different benchmark datasets. Our findings highlight that Mamba-based models, $\mathcal{DG}$-Mamba and $\mathcal{GDG}$-Mamba, fairly outperformed transformer-based approaches for temporal link prediction tasks. Implementing selective state-space models enables significant reductions in computational complexity as these models incorporate more hardware-aware mechanisms and are more hardware-optimized. One interesting finding in this paper reveals how Mamba exhibits attention-like behavior, showing how state-space models effectively capture temporal dependencies. We make this discovery after analysis of the state matrix $A$ in Mamba to formulate the hidden attention mechanism. Moreover, Mamba-based architectures not only overcome the quadratic complexity of traditional transformers but also provide an effective solution for long-range dependencies in dynamic graphs. Based on our evaluation, we can infer that $\mathcal{GDG}$-Mamba, by incorporating Graph Isomorphism Network Edge (GINE) convolutions in $\mathcal{DG}$-Mamba, gives better results because it enhances spatial representation learning by leveraging node-level and edge-level features.

Our research highlights the potential of Mamba-based architectures for scaling temporal graph modeling to real-world applications, such as biological systems modeling, financial forecasting, or social network analysis. Through these advances, this study contributes a robust foundation for developing efficient and scalable solutions in temporal graph representation learning. Future work could explore hybrid architectures and the extension of these methods to other graph-based tasks, such as anomaly detection and clustering.

\section*{Acknowledgements}
We would like to acknowledge support from the DOE SEA-CROGS project (DE-SC0023191) and the AFOSR project (FA9550-24-1-0231).


\begin{thebibliography}{10}
\expandafter\ifx\csname url\endcsname\relax
  \def\url#1{\texttt{#1}}\fi
\expandafter\ifx\csname urlprefix\endcsname\relax\def\urlprefix{URL }\fi
\expandafter\ifx\csname href\endcsname\relax
  \def\href#1#2{#2} \def\path#1{#1}\fi

\bibitem{kipf2016semi}
T.~N. Kipf, M.~Welling, Semi-supervised classification with graph convolutional networks, arXiv preprint arXiv:1609.02907 (2016).

\bibitem{gilmer2017neural}
J.~Gilmer, S.~S. Schoenholz, P.~F. Riley, O.~Vinyals, G.~E. Dahl, Neural message passing for quantum chemistry, in: International conference on machine learning, PMLR, 2017, pp. 1263--1272.

\bibitem{velivckovic2017graph}
P.~Veli{\v{c}}kovi{\'c}, G.~Cucurull, A.~Casanova, A.~Romero, P.~Lio, Y.~Bengio, Graph attention networks, arXiv preprint arXiv:1710.10903 (2017).

\bibitem{yun2019graph}
S.~Yun, M.~Jeong, R.~Kim, J.~Kang, H.~J. Kim, Graph transformer networks, Advances in neural information processing systems 32 (2019).

\bibitem{gou2023discriminative}
J.~Gou, X.~Yuan, Y.~Xue, L.~Du, J.~Yu, S.~Xia, Y.~Zhang, Discriminative and geometry-preserving adaptive graph embedding for dimensionality reduction, Neural Networks 157 (2023) 364--376.

\bibitem{trivedi2019dyrep}
R.~Trivedi, M.~Farajtabar, P.~Biswal, H.~Zha, {DyRep}: Learning representations over dynamic graphs, in: International conference on learning representations, 2019.

\bibitem{kumar2019predicting}
S.~Kumar, X.~Zhang, J.~Leskovec, Predicting dynamic embedding trajectory in temporal interaction networks, in: Proceedings of the 25th ACM SIGKDD international conference on knowledge discovery \& data mining, 2019, pp. 1269--1278.

\bibitem{rossi2020temporal}
E.~Rossi, B.~Chamberlain, F.~Frasca, D.~Eynard, F.~Monti, M.~Bronstein, Temporal graph networks for deep learning on dynamic graphs, arXiv preprint arXiv:2006.10637 (2020).

\bibitem{goyal2018dyngem}
P.~Goyal, N.~Kamra, X.~He, Y.~Liu, {DynGEM}: Deep embedding method for dynamic graphs, arXiv preprint arXiv:1805.11273 (2018).

\bibitem{goyal2018dynamicgem}
P.~Goyal, S.~R. Chhetri, N.~Mehrabi, E.~Ferrara, A.~Canedo, {DynamicGEM: A library for dynamic graph embedding methods}, arXiv preprint arXiv:1811.10734 (2018).

\bibitem{pareja2020evolvegcn}
A.~Pareja, G.~Domeniconi, J.~Chen, T.~Ma, T.~Suzumura, H.~Kanezashi, T.~Kaler, T.~Schardl, C.~Leiserson, {EvolveGCN}: Evolving graph convolutional networks for dynamic graphs, in: Proceedings of the AAAI conference on artificial intelligence, Vol.~34, 2020, pp. 5363--5370.

\bibitem{xu2022dyng2g}
M.~Xu, A.~V. Singh, G.~E. Karniadakis, {DynG2G}: An efficient stochastic graph embedding method for temporal graphs, IEEE Transactions on Neural Networks and Learning Systems 35~(1) (2022) 985--998.

\bibitem{you2022roland}
J.~You, T.~Du, J.~Leskovec, {ROLAND}: graph learning framework for dynamic graphs, in: Proceedings of the 28th ACM SIGKDD conference on knowledge discovery and data mining, 2022, pp. 2358--2366.

\bibitem{sharma2023temporal}
K.~Sharma, R.~Trivedi, R.~Sridhar, S.~Kumar, Temporal dynamics-aware adversarial attacks on discrete-time dynamic graph models, in: Proceedings of the 29th ACM SIGKDD Conference on Knowledge Discovery and Data Mining, 2023, pp. 2023--2035.

\bibitem{vaswani2017attention}
A.~Vaswani, N.~Shazeer, N.~Parmar, J.~Uszkoreit, L.~Jones, A.~N. Gomez, {\L}.~Kaiser, I.~Polosukhin, Attention is all you need, Advances in neural information processing systems 30 (2017).

\bibitem{varghese2024transformerg2g}
A.~J. Varghese, A.~Bora, M.~Xu, G.~E. Karniadakis, Transformer{G2G}: Adaptive time-stepping for learning temporal graph embeddings using transformers, Neural Networks 172 (2024) 106086.

\bibitem{liu2021anomaly}
Y.~Liu, S.~Pan, Y.~G. Wang, F.~Xiong, L.~Wang, Q.~Chen, V.~C. Lee, Anomaly detection in dynamic graphs via transformer, IEEE Transactions on Knowledge and Data Engineering 35~(12) (2021) 12081--12094.

\bibitem{huo2023hierarchical}
G.~Huo, Y.~Zhang, B.~Wang, J.~Gao, Y.~Hu, B.~Yin, Hierarchical spatio--temporal graph convolutional networks and transformer network for traffic flow forecasting, IEEE Transactions on Intelligent Transportation Systems 24~(4) (2023) 3855--3867.

\bibitem{jiang2023pdformer}
J.~Jiang, C.~Han, W.~X. Zhao, J.~Wang, {PDFormer: Propagation delay-aware dynamic long-range transformer for traffic flow prediction}, in: Proceedings of the AAAI conference on artificial intelligence, Vol.~37, 2023, pp. 4365--4373.

\bibitem{lezmi2023time}
E.~Lezmi, J.~Xu, Time series forecasting with transformer models and application to asset management, Available at SSRN 4375798 (2023).

\bibitem{wang2024dynamic}
J.~Wang, Z.~Sun, C.~Yuan, W.~Li, A.-A. Liu, Z.~Wei, B.~Yin, Dynamic graphs attention for ocean variable forecasting, Engineering Applications of Artificial Intelligence 133 (2024) 108187.

\bibitem{hofmeister2024dynamic}
M.~Hofmeister, K.~F. Lee, Y.-K. Tsai, M.~M{\"u}ller, K.~Nagarajan, S.~Mosbach, J.~Akroyd, M.~Kraft, Dynamic control of district heating networks with integrated emission modelling: A dynamic knowledge graph approach, Energy and AI 17 (2024) 100376.

\bibitem{ren2023dynamic}
S.~Ren, X.~Pan, W.~Zhao, B.~Nie, B.~Han, Dynamic graph transformer for {3D} object detection, Knowledge-Based Systems 259 (2023) 110085.

\bibitem{gu2021efficiently}
A.~Gu, K.~Goel, C.~R{\'e}, Efficiently modeling long sequences with structured state spaces, arXiv preprint arXiv:2111.00396 (2021).

\bibitem{gu2023mamba}
A.~Gu, T.~Dao, Mamba: Linear-time sequence modeling with selective state spaces, arXiv preprint arXiv:2312.00752 (2023).

\bibitem{longa2023graph}
A.~Longa, V.~Lachi, G.~Santin, M.~Bianchini, B.~Lepri, P.~Lio, F.~Scarselli, A.~Passerini, Graph neural networks for temporal graphs: State of the art, open challenges, and opportunities, arXiv preprint arXiv:2302.01018 (2023).

\bibitem{li2024state}
J.~Li, R.~Wu, X.~Jin, B.~Ma, L.~Chen, Z.~Zheng, State space models on temporal graphs: A first-principles study, arXiv preprint arXiv:2406.00943 (2024).

\bibitem{holme2012temporal}
P.~Holme, J.~Saram{\"a}ki, Temporal networks, Physics reports 519~(3) (2012) 97--125.

\bibitem{kazemi2020representation}
S.~M. Kazemi, R.~Goel, K.~Jain, I.~Kobyzev, A.~Sethi, P.~Forsyth, P.~Poupart, Representation learning for dynamic graphs: A survey, Journal of Machine Learning Research 21~(70) (2020) 1--73.

\bibitem{xu2020inductive}
D.~Xu, C.~Ruan, E.~Korpeoglu, S.~Kumar, K.~Achan, Inductive representation learning on temporal graphs, in: International Conference on Learning Representations, 2020.

\bibitem{singer2019node}
U.~Singer, I.~Guy, K.~Radinsky, Node embedding over temporal graphs, arXiv preprint arXiv:1903.08889 (2019).

\bibitem{seo2018structured}
Y.~Seo, M.~Defferrard, P.~Vandergheynst, X.~Bresson, Structured sequence modeling with graph convolutional recurrent networks, in: International Conference on Neural Information Processing, Springer, 2018, pp. 362--373.

\bibitem{patro2024mamba}
B.~N. Patro, V.~S. Agneeswaran, Mamba-360: Survey of state space models as transformer alternative for long sequence modelling: Methods, applications, and challenges, arXiv preprint arXiv:2404.16112 (2024).

\bibitem{behrouz2024graph}
A.~Behrouz, F.~Hashemi, Graph mamba: Towards learning on graphs with state space models, arXiv preprint arXiv:2402.08678 (2024).

\bibitem{bojchevski2018deep}
A.~Bojchevski, S.~G{\"u}nnemann, Deep gaussian embedding of graphs: Unsupervised inductive learning via ranking, in: International Conference on Learning Representations, 2018, pp. 1--13.

\bibitem{optuna_2019}
T.~Akiba, S.~Sano, T.~Yanase, T.~Ohta, M.~Koyama, Optuna: A next-generation hyperparameter optimization framework, in: Proceedings of the 25th {ACM} {SIGKDD} International Conference on Knowledge Discovery and Data Mining, 2019.

\bibitem{goyal2020dyngraph2vec}
P.~Goyal, S.~R. Chhetri, A.~Canedo, dyngraph2vec: Capturing network dynamics using dynamic graph representation learning, Knowledge-Based Systems 187 (2020) 104816.

\bibitem{hidden_mamba}
A.~Ali, I.~Zimerman, L.~Wolf, The hidden attention of mamba models, arXiv preprint arXiv:2403.01590v2 (2024).

\end{thebibliography}


\newpage
\end{document}